\documentclass[twoside,11pt]{article}

\usepackage{jmlr2e}
\usepackage{microtype}
\usepackage{subfigure}
\usepackage{booktabs} 
\usepackage{amsmath}
\usepackage{url}
\usepackage{amsfonts}       
\usepackage{nicefrac}       
\usepackage{xcolor}
\usepackage[export]{adjustbox}

\usepackage{bm}

\newcommand{\eop}{\hfill\ensuremath{\square}}
\def\BE{\begin{equation}}
\def\EE{\end{equation}}
\def\BEA{\begin{eqnarray}}
\def\EEA{\end{eqnarray}}

\ShortHeadings{A Bayes-Optimal View on Adversarial Examples}{Richardson and Weiss}
\firstpageno{1}

\begin{document}

\title{A Bayes-Optimal View on Adversarial Examples}

\author{\name Eitan Richardson \email eitanrich@cs.huji.ac.il \\
       \addr School of Computer Science and Engineering\\
       The Hebrew University of Jerusalem\\
       Jerusalem, Israel
       \AND
       \name Yair Weiss \email yweiss@cs.huji.ac.il \\
       \addr School of Computer Science and Engineering\\
       The Hebrew University of Jerusalem\\
       Jerusalem, Israel}

\editor{Rob Fergus}

\maketitle

\begin{abstract}
Since the discovery of adversarial examples -- the ability to fool modern CNN classifiers with tiny perturbations of the input, there has been much discussion whether they are a ``bug" that is specific to current neural architectures and training methods or an inevitable ``feature" of high dimensional geometry. In this paper, we argue for examining adversarial examples from the perspective of \emph{Bayes-Optimal classification}. We construct realistic image datasets for which the Bayes-Optimal classifier can be efficiently computed and derive analytic conditions on the distributions under which these classifiers are \emph{provably robust against any adversarial attack even in high dimensions}. Our results show that even when these ``gold standard'' optimal classifiers are robust, CNNs trained on the same datasets consistently learn a vulnerable classifier, indicating that adversarial examples are often an avoidable ``bug''. We further show that RBF SVMs trained on the same data consistently learn a robust classifier. The same trend is observed in experiments with real images in different datasets.
\end{abstract}

\begin{keywords}
  Adversarial Examples, Bayes Optimal, Generative Models, CNN, SVM
\end{keywords}

\section{Introduction}

\begin{figure}
    \centering
    \begin{small}
    \begin{tabular}{ccccc}
    \multicolumn{1}{c|}{Original ``male"} & \multicolumn{4}{c}{Adversarial ``female"}\\
    \multicolumn{1}{c|}{\includegraphics[width=0.16\textwidth]{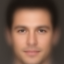}} &
    \includegraphics[width=0.16\textwidth]{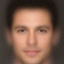} &
    \includegraphics[width=0.16\textwidth]{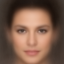} &
    \includegraphics[width=0.16\textwidth]{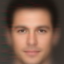} &
    \includegraphics[width=0.16\textwidth]{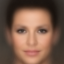} \\
    \multicolumn{1}{c|}{} & Asymmetric dataset & Symmetric dataset & Symmetric dataset & Symmetric dataset \\
    \multicolumn{1}{c|}{} & Bayes-Optimal & Bayes-Optimal & CNN & SVM (RBF) \\
    (a) & (b) & (c) & (d) & (e) \rule{0pt}{1.5em}
    \end{tabular}
    \end{small}
    \caption{We present realistic image datasets for which the Bayes-Optimal classifier can be calculated efficiently and derive analytic conditions on when the optimal classifier will be robust or vulnerable to adversarial examples. When the data distribution satisfies certain asymmetries, the Bayes-Optimal classifier  is vulnerable (b), but when the distribution is symmetric, the optimal classifier is
    robust and adversarial attacks become perceptually meaningful (c). Our experiments with these datasets show that CNN training consistently learns a vulnerable classifier (d) even when the optimal classifier is robust, while large-margin methods often succeed (e). }
    \label{teaser-fig}
\end{figure}

Perhaps the most intriguing property of modern machine learning methods is their susceptibility to adversarial examples~\citep{SzegedyZSBEGF13}: for many powerful classifiers it is possible to perturb the input by an imperceptible amount and change the decision of the classifier. While adversarial examples were most famously reported for CNN classifiers~\citep{SzegedyZSBEGF13}, subsequent research has shown that other classifiers can also fall prey to similar attacks~\citep{Goodfellow2018}. Attempts to make classifiers robust to these attacks have generated a tremendous amount of interest (e.g.~\citealt{Schott2018a} and references within).

As a first step towards solving the problem, many authors have attempted to understand the source of the failure~\citep{Goodfellow2018,SzegedyZSBEGF13,tilting,FawziFF18,shamir}. Broadly speaking, existing explanations fall into two groups (see section~\ref{sec:related} for a more detailed discussion of related work).  One approach argues that adversarial vulnerability is in some sense inevitable: either due to the geometry of high dimensions (e.g.~\citealt{Goodfellow2018,GoodfellowSS14,shamir}) or due to fundamental limitations on robustness of classifiers trained from finite data~\citep{Schmidt2018}. Another approach views adversarial vulnerability as a ``bug" of CNNs and current training methods, which can be avoided by other architectures or training protocols (e.g.~\citealt{tilting,nakkiran2019a,lyu2020gradient,Schott2018a}). 

One reason for the existence of many conflicting explanations may be the difficulty of analyzing adversarial vulnerability in a realistic yet tractable setting. In this paper, we provide such a setting.  We construct realistic image datasets for which the Bayes-Optimal classifier can be efficiently computed and analyze the vulnerability of the optimal classifier. We derive analytic conditions on the distributions where {\em even the optimal classifier will be vulnerable} and other conditions where {\em the optimal classifier will be provably robust}. Figure~\ref{teaser-fig}a\nobreakdash-c shows an example: our synthetic dataset of ``male" and ``female" face images. In the first, ``asymmetric" distribution, the Bayes-Optimal classifier is  vulnerable and an imperceptible perturbation is sufficient to change a ``male" face (Figure~\ref{teaser-fig}a) to one that would be classified as ``female" (Figure~\ref{teaser-fig}b). In the second, ``symmetric" dataset, fooling the optimal classifier requires making large, perceptually meaningful changes (Figure~\ref{teaser-fig}c).

By training different classifiers on the ``symmetric'' datasets, in which the optimal classifiers are robust, we can disentangle the possible sources of vulnerability and avoid the accuracy-robustness tradeoff that may occur in commonly used datasets. Our results show that even when the optimal classifier is robust, standard CNN training consistently learns a vulnerable classifier (Figure~\ref{teaser-fig}d).
At the same time, for exactly the same training data, RBF SVMs consistently learn a robust classifier (Figure~\ref{teaser-fig}e).
Our results suggest that in many realistic settings, adversarial vulnerability is not an unavoidable property of learning in high dimensions but rather a direct result of suboptimal training methods used in current practice. 

The paper is organized as follows: In section~\ref{sec:realistic-datasets} we present our method for constructing realistic image datasets that are equipped with computable Bayes-optimal classifiers. We then show, in sections~\ref{sec:conditions}, under which conditions these optimal classifiers are provably robust against adversarial attacks. Section~\ref{sec:experiments} contains our experimental evaluation of the vulnerability of different classifiers trained on these datasets. Additional related work is discussed in section~\ref{sec:related}.

\section{A Realistic Tractable Setting for Analysing Adversarial Vulnerability}

This section describes our method for generating realistic image datasets with computable Bayes-Optimal classifiers that are provably robust to all adversarial attacks.

\begin{figure}
  \centerline{
    \begin{tabular}{cc}
     \raisebox{-0.5\totalheight}{\includegraphics[width=0.32\textwidth]{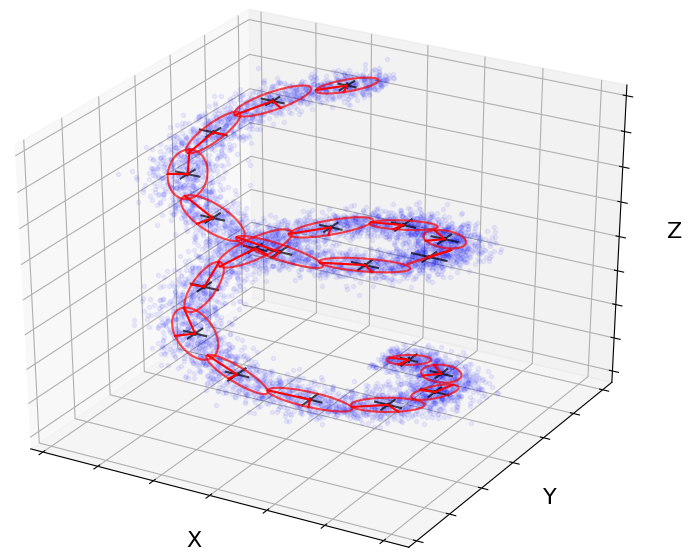}} & 
    \scriptsize
    \begin{tabular}{cc}
     Male: & Female:     \\
     \includegraphics[trim=0 0 128px 0,clip, width=0.32\textwidth]{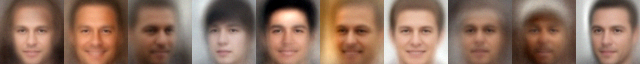} &
     \includegraphics[trim=0 0 128px 0,clip,width=0.32\textwidth]{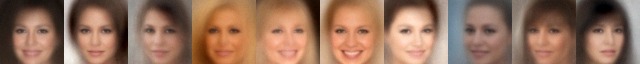} \\
     Smiling: & Not Smiling: \\
     \includegraphics[trim=0 0 128px 0,clip,width=0.32\textwidth]{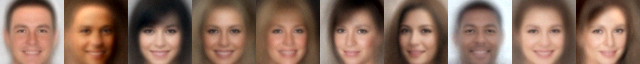} &  \includegraphics[trim=0 0 128px 0,clip,width=0.32\textwidth]{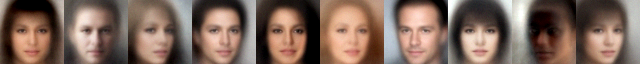} \\
     Eyeglasses: & No Eyeglasses: \\
     \includegraphics[trim=0 0 128px 0,clip,width=0.32\textwidth]{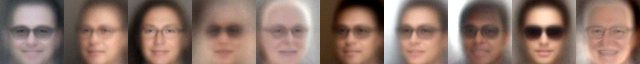} &  \includegraphics[trim=0 0 128px 0,clip,width=0.32\textwidth]{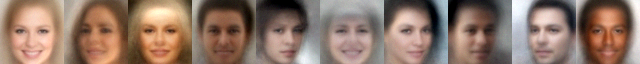} \\
     0: & 6: \\
     \includegraphics[trim=0 0 56px 0,clip,width=0.32\textwidth]{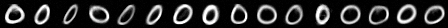} &  
     \includegraphics[trim=0 0 56px 0,clip,width=0.32\textwidth]{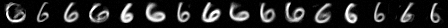} \\
     2: & 7: \\
     \includegraphics[trim=0 0 56px 0,clip,width=0.32\textwidth]{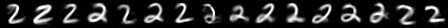} &  
     \includegraphics[trim=0 0 56px 0,clip,width=0.32\textwidth]{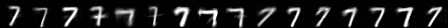}
    \end{tabular}
    \end{tabular}
  }
  \caption[]{Left: A Mixture of Factor Analyzers (MFA) model for toy data in $\mathbb{R}^3$ (sampled 3D helical surface -- blue points). Each component is a Gaussian on a learned 2-dimensional hyperplane with added axis-aligned noise. Right: Samples from our realistic MFA datasets.}
\label{fig:samples}
\end{figure}

\subsection{Image datasets with computable Bayes-Optimal classifiers}
\label{sec:realistic-datasets}

We focus on a two class classification problem and  denote by $p_1(x)$, $p_2(x)$  the distribution of the two classes. We assume that $p_1(x),p_2(x)$ are known and that the two classes have equal priors, so the {\em Bayes-Optimal} classifier simply classifies $x$ as belonging to class $1$ if $p_1(x)>p_2(x)$ and to class $2$ otherwise. It is well known that this classification rule is optimal and no other classification rule can achieve higher accuracy (assuming of course that $p_1(x),p_2(x)$ are correct)~\citep{duda1973pattern}. Several recent papers ~\citep{Schmidt2018,ilyas2019adversarial} have analyzed adversarial examples in this setting, but only when $p_1(x),p_2(x)$ are both Gaussians with the same covariance, so that the optimal classifier is linear. Real image classification problems are of course very different from this simplified setting.

In our approach, we assume that images that belong to a single class form a {\em nonlinear, low-dimensional manifold} in pixel space. A standard way to model such manifolds is to use a mixture of Factor Analyzers (MFA) model~\citep{ghahramani1996algorithm}. The model is based on the observation that {\em locally} the manifold can be represented by a Gaussian whose covariance matrix is a sum of a low-rank matrix (representing the covariance on the manifold) and a diagonal matrix (representing the covariance off the manifold, e.g.\ due to sensor noise). Figure~\ref{fig:samples} (left) shows an example of a nonlinear 2D manifold in three dimensions, and its representation using an MFA model. 

To create synthetic image datasets for classification, we start with a labeled training set  with two classes. We then train separate MFA models $p_1(x)$, $p_2(x)$ on images from the two classes using the algorithm (and code) provided by \citet{RichardsonW18}.  We now create a new training set by sampling images from the two models, and similarly a new test set. Since these datasets were created by sampling from a known model, we can calculate the Bayes-Optimal classifier. At the same time, the images are realistic and MFA models have been shown to capture much of the variability of the images in the original data~\citep{RichardsonW18}.

We created 12 such datasets of faces (based on the CelebA dataset~\citealt{liu2015faceattributes}) and 3 datasets of digits (based on MNIST)\footnote{The datasets and models will be made publicly available after publication.}.  Figure~\ref{fig:samples} (right) shows samples from five such datasets that correspond to five binary classification problems: Male vs.\ Female, Smiling vs.\ Not-smiling, Eyeglasses vs. No-eyeglasses, 0 vs.\ 6, and 2 vs.\ 7.  While these samples are typically somewhat blurred, it can be seen  they are realistic and highly variable. In fact, in many real world applications one needs to classify somewhat blurry images (e.g.\ analyzing faces in surveillance videos). We again emphasize that for all of these datasets, we can efficiently calculate the Bayes-Optimal classifier. 

The advantage of using a MFA model over other generative models such as VAEs or GANs~\citep{kingma2013auto,gulrajani2017improved} is that the log likelihood of any image can be calculated efficiently.  Since the data were generated by the assumed distributions, this classifier is Bayes-Optimal. Indeed in all datasets we created, the accuracy of the classifier was close to 100\%.  

We now show under what conditions these optimal classifiers are \emph{provably robust to all adversarial attacks}.

\subsection{Provably robust or vulnerable optimal classifiers }
\label{sec:conditions}

\begin{figure}
  \centerline{
    \begin{tabular}{cccc}
     Spherical Gaussians & Elliptical Gaussians & Discrete & Mixtures\\
      \includegraphics[width=0.22\columnwidth]{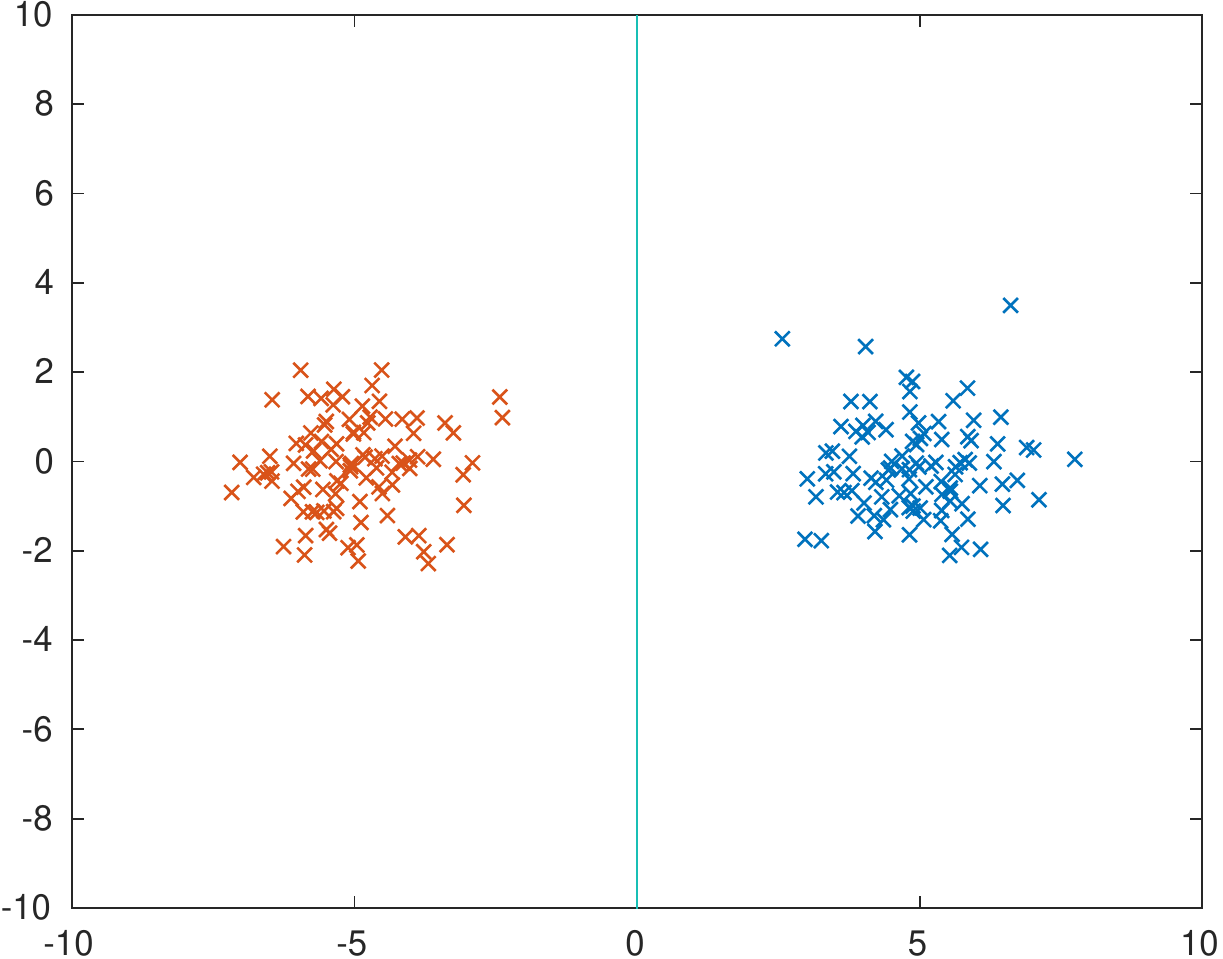}
      &
      \includegraphics[width=0.24\columnwidth]{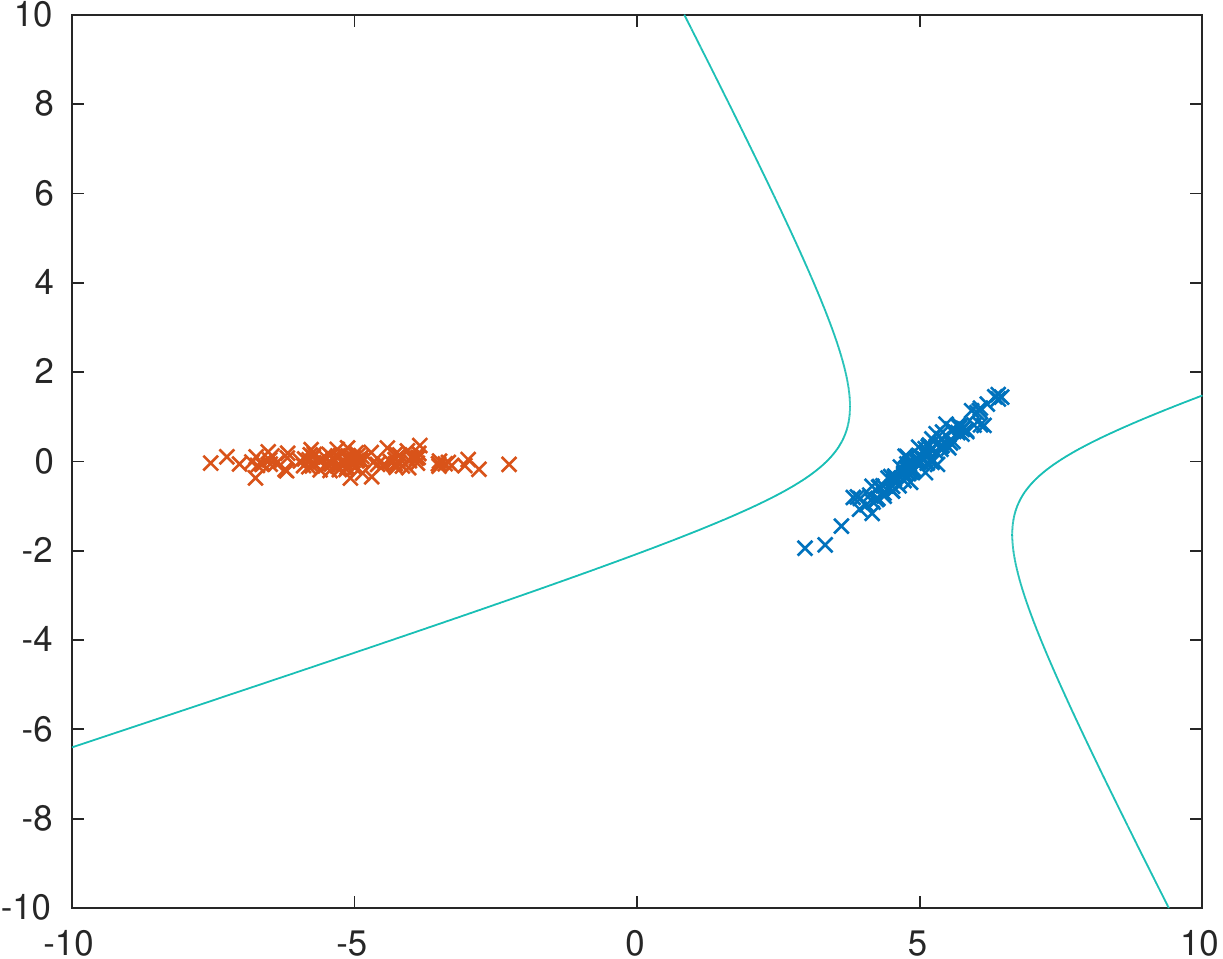} &
      \includegraphics[width=0.24\columnwidth]{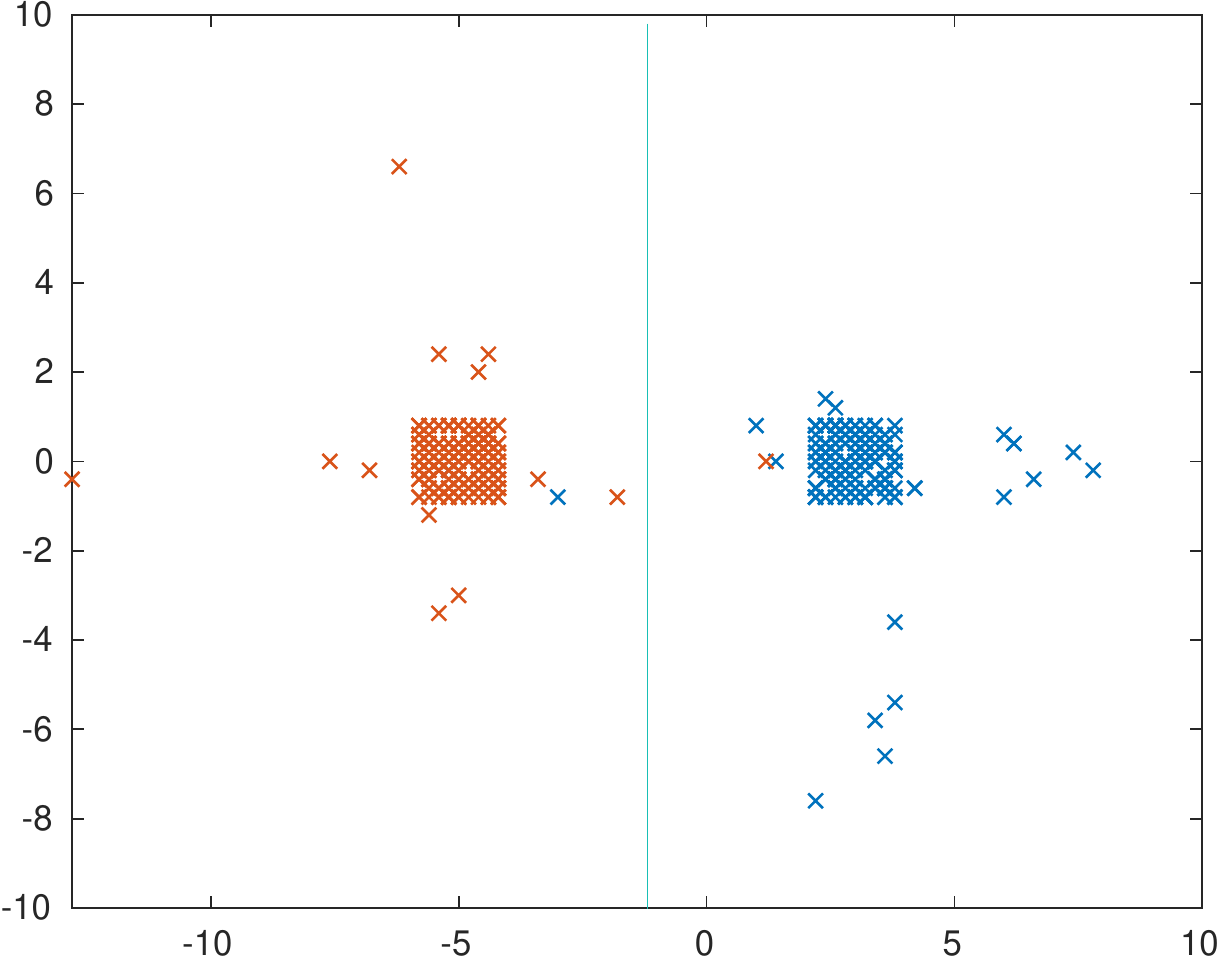} &
      \includegraphics[width=0.24\columnwidth]{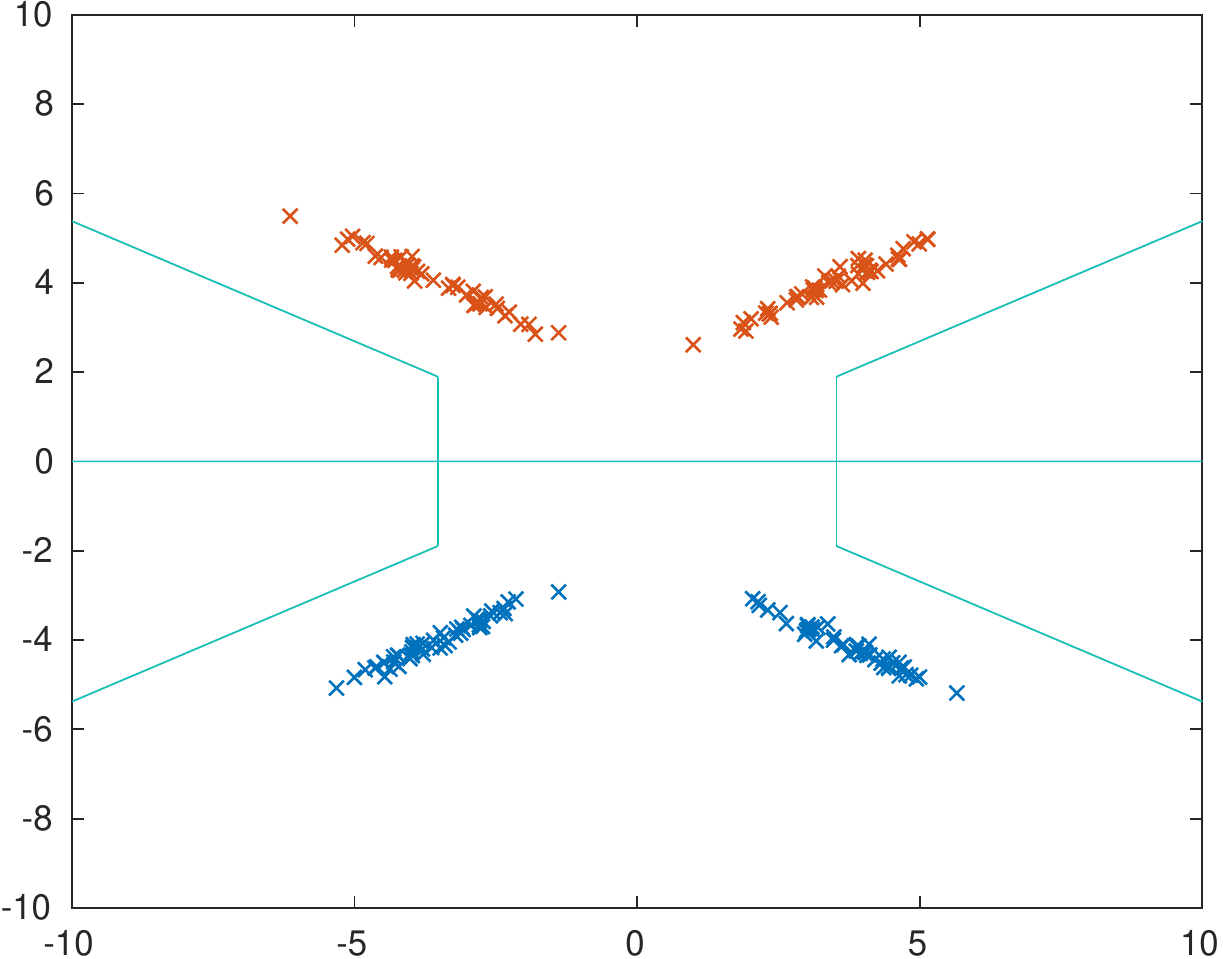} \\
(a) & (b) & (c) & (d) \\
      \includegraphics[width=0.24\columnwidth]{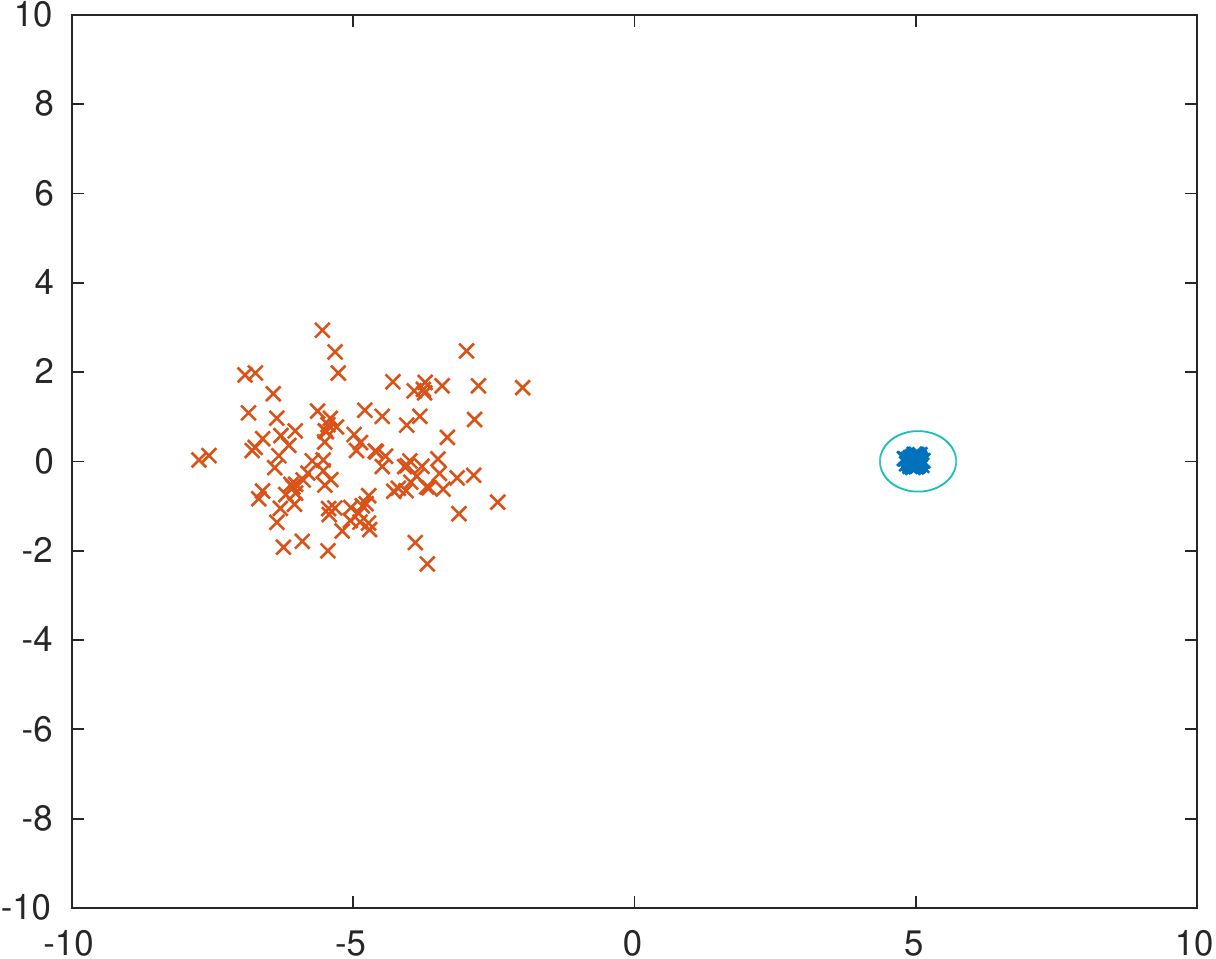}
      &
      \includegraphics[width=0.24\columnwidth]{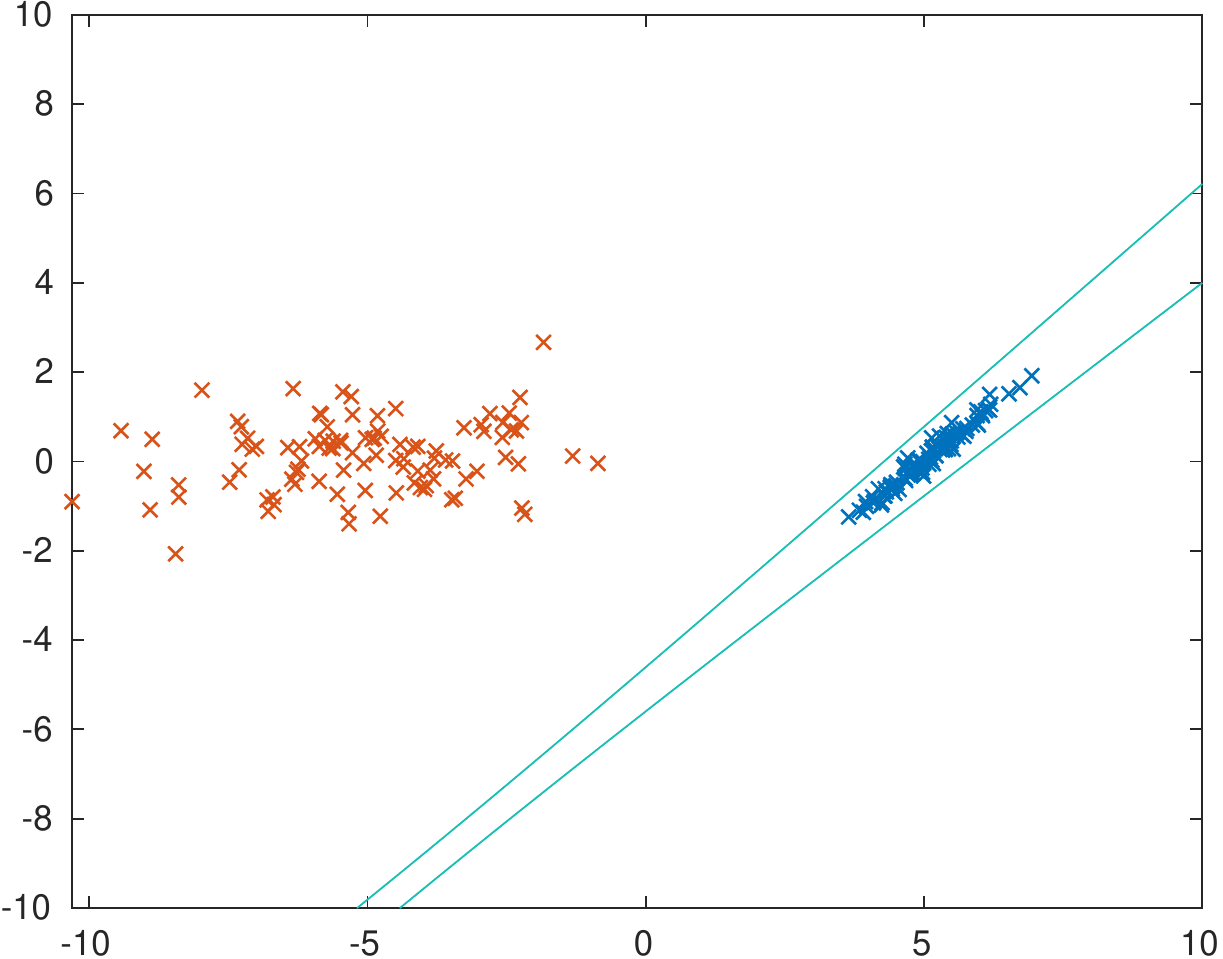} &
      \includegraphics[width=0.24\columnwidth]{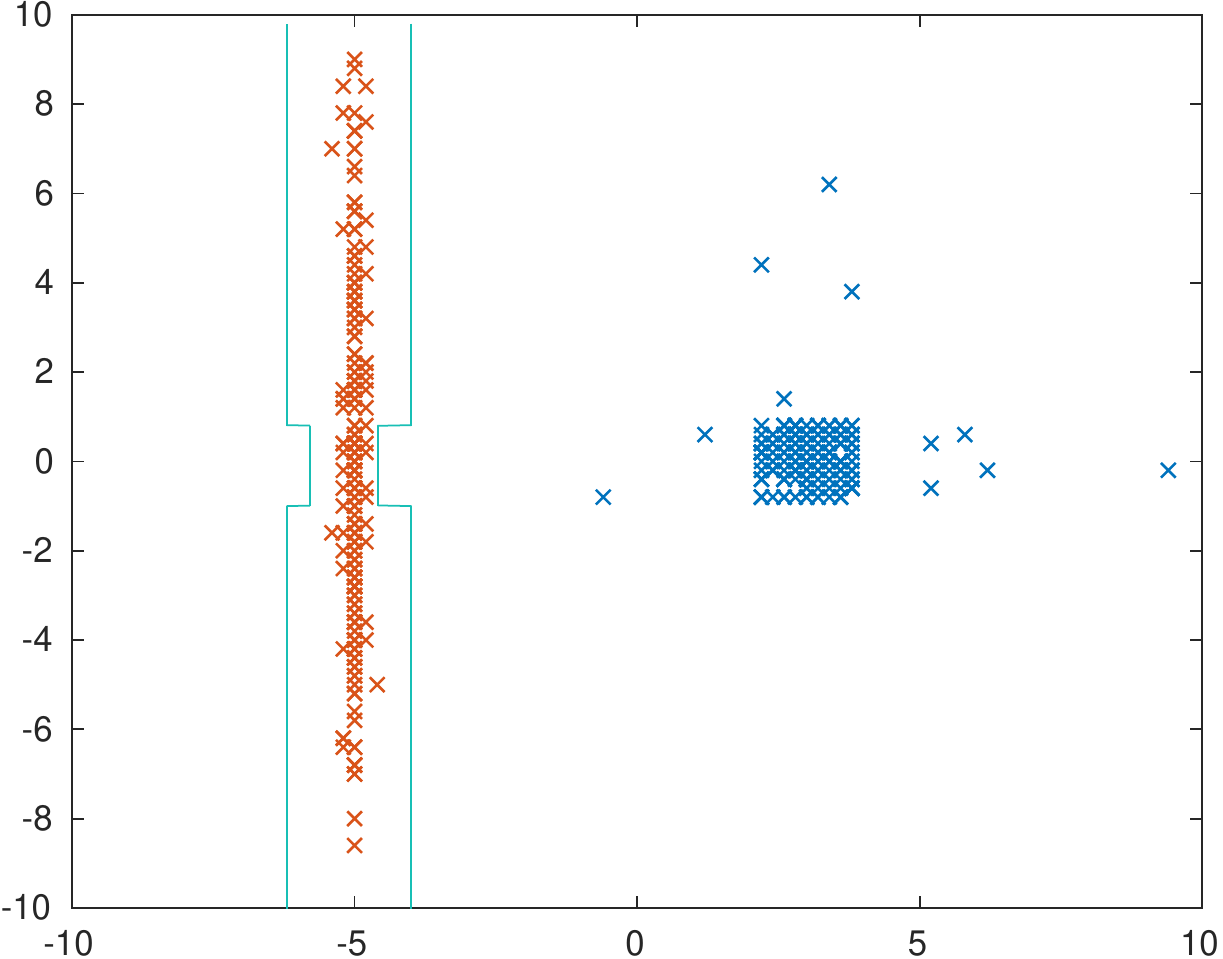} &
      \includegraphics[width=0.24\columnwidth]{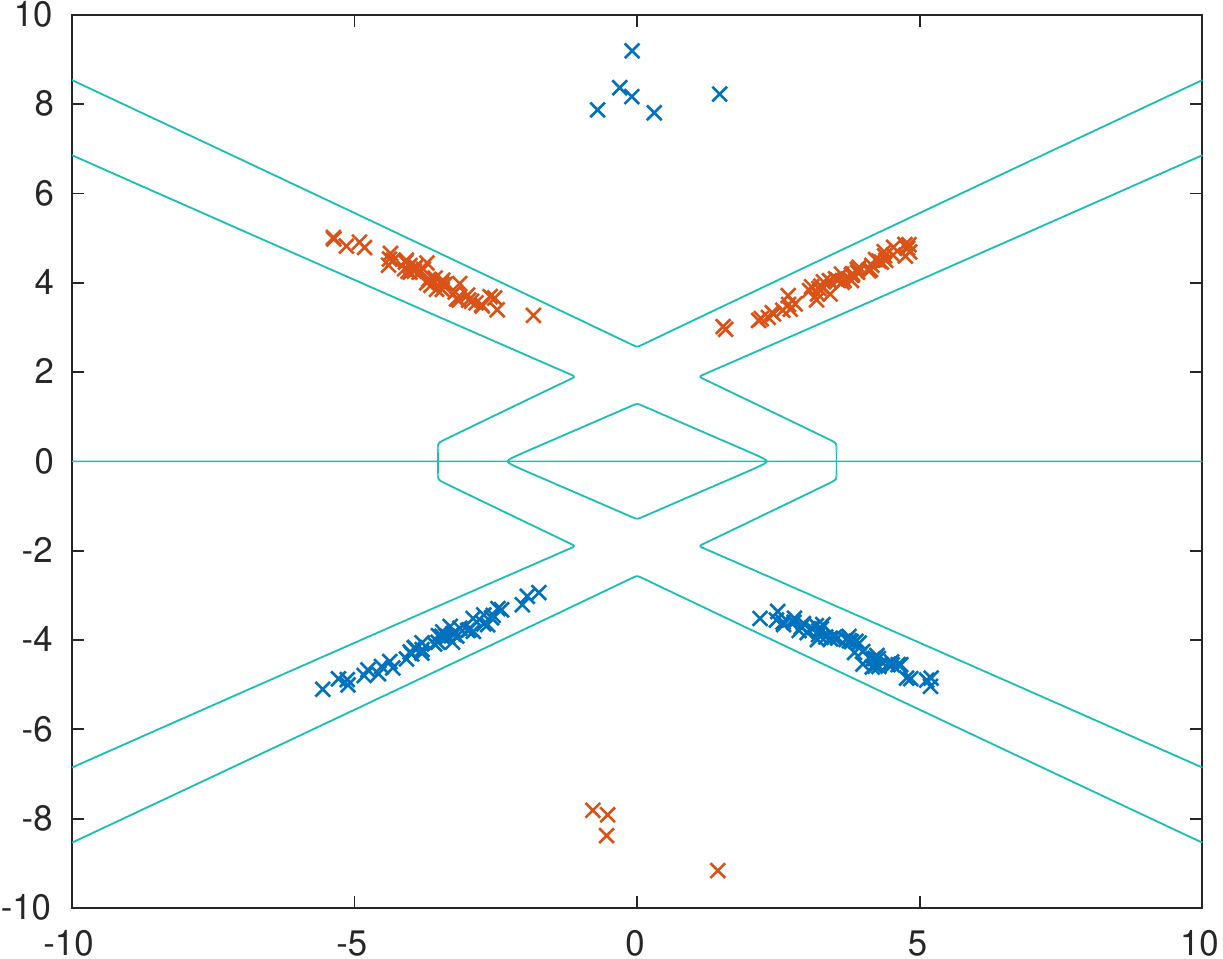} \\
(e) & (f) & (g) & (h)
    \end{tabular}
  }
  \caption[]{Top: examples of 2D distribution where the optimal decision boundary is far from the datapoints. Bottom: distribution where the optimal boundary is close to many of the datapoints.}
\label{figure-toy}
\end{figure}

A textbook example of Bayes-Optimal classification is when both classes are generated using Gaussians with the same spherical covariance, in which case the optimal classifier is a linear discriminant that is orthogonal to the difference between the two means (Figure~\ref{figure-toy}a). In this case, if the distance between the two means is large relative to the covariance, then almost all points are far from the decision boundary and so any adversarial attack which only makes small changes to the input will typically fail. But as shown in the bottom of Figure~\ref{figure-toy} there are other examples where the decision boundary is close to many of the datapoints and an adversarial attack which only makes small changes to the input will often succeed\footnote{We deem an attack successful if it causes the classifier to change its prediction. Other definitions exist \citep{diochnos2018adversarial}, but prediction-change is the most commonly used and practical one.}. What distinguishes these two cases? The following three theorems (whose proofs are given in the appendix) summarize our results. 

{\bf Theorem 1: \emph{Symmetric} isotropic Gaussian.} If $p_1(x)=N(\mu,\sigma^2 I )$, $p_2(x)={N(\mu+d,\sigma^2I)}$ and $\sigma<<\|d\|$ then for almost all points the Bayes-Optimal classifier is robust to any perturbation smaller than $\|d\|/2$.

{\bf Theorem 2:  \emph{Symmetric} MFA.} If $p_1(x),p_2(x)$ are MFA models, i.e. a GMM where each Gaussian is of the form $N(\mu_i,\Sigma_i)$ and $\Sigma_i$ is a 
covariance that is a sum of a low-rank matrix plus a diagonal matrix
$\Sigma_i=A_i A_i^T + \sigma^2 I$ and all $A_i$ have the same singular values.  Let $\bm{d}$ be the minimal distance between any two subspaces of different manifolds\footnote{Two low-dimensional subspaces in high dimensions will  almost always not intersect.}:
\[
\bm{d}= \min_{i,j,z_1,z_2} \| \mu_i + A_i z_1 - (\mu_j + A_j z_2)\|
\]
As $\sigma \rightarrow 0$ then for almost all points the Bayes-Optimal classifier is robust to any perturbation smaller than $\bm{d}/2$.

Theorems 1-2 referred to the \emph{symmetric} case, in which the optimal classifier is \emph{robust}. The next theorem refers to the \emph{asymmetric} case, in which the optimal classifier is \emph{vulnerable}:

{\bf Lemma 1: \emph{Asymmetric} Gaussian.}  Assume $p_1,p_2$ are Gaussian distributions. Let $v$ be a direction of minimal variance under $\Sigma_1$: $v=\arg\min v'^T \Sigma_1 v'$. Let $\sigma_1^2$ be the variance projecting $x$
onto that direction when $x$ comes from class $1$.  
If $\sigma_1 \rightarrow 0$ and $\Sigma_2$ is full rank then almost any point in class $1$ is arbitrarily close to the optimal decision surface. 

{\bf Theorem 3: \emph{Asymmetric} GMM.} Assume that both  $p_1(x)$ and $p_2(x)$ are Gaussian Mixture Models. If for every Gaussian in one class there exists a component in the other class so that they satisfy the \emph{asymmetry} conditions of Lemma 1, then almost any point will be arbitrarily close to the optimal decision boundary. 

\begin{figure}
    \centering
    \begin{scriptsize}
    \begin{tabular}{c}
      Samples from $p_1$: \\
      \includegraphics[width=0.6\columnwidth]{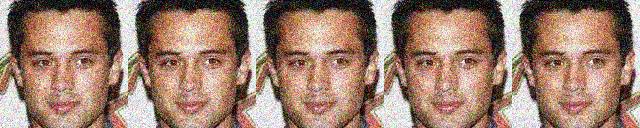} \\
      Samples from $p_2$: \\
      \includegraphics[width=0.6\columnwidth]{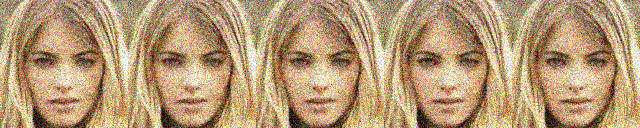} \\
      \\
    \begin{tabular}{rcc|cc}
         & & Perturbation & & Perturbation \\
         Original: & \includegraphics[height=16mm]{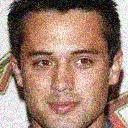} & & 
         \includegraphics[height=16mm]{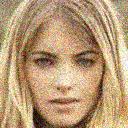} & \\
         Optimal: & \includegraphics[height=16mm]{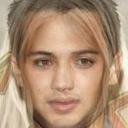} &
         \includegraphics[height=16mm]{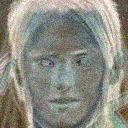} &
         \includegraphics[height=16mm]{./Figs/twoface/boundary.jpg} &
         \includegraphics[height=16mm]{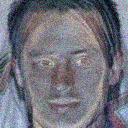} \\
         CNN: & \includegraphics[height=16mm]{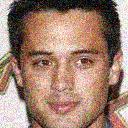} & 
         \includegraphics[height=16mm]{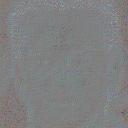} &
         \includegraphics[height=16mm]{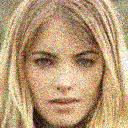} & 
         \includegraphics[height=16mm]{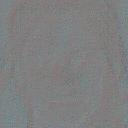} \\
    \end{tabular}{}
    \end{tabular}
    \end{scriptsize}
    \caption{Top: Samples from our ``noisy two-face'' toy data (each class contains noisy versions of the same face image). Bottom: Attacking test samples. The Bayes-Optimal model is robust (middle row) while a CNN trained on this simple data is vulnerable (bottom row).}
    \label{fig:noisy-two-face}
\end{figure}

Perhaps the most striking aspect of our three theorems is that they do not depend on the dimensions of the inputs $x$. Thus Theorem 1 guarantees that for two well-separated Gaussians, the Bayes-Optimal classifier is robust {\em for arbitrarily large dimensions.}
 Figure~\ref{fig:noisy-two-face} illustrates 
 Theorem 1 for the toy case where the two Gaussian means are two specific face images and the inputs lie in $R^{49,152}$ . As can be seen, the optimal classifier is robust -- in order to fool it, the adversary needs to make large, perceptual-meaningful changes to the image. Specifically the decision boundary,  like in Figure~\ref{figure-toy}a is a plane that is evenly-distanced to the two means. Samples drawn from these distributions can then be used as training sets for evaluating the robustness of, for example, CNNs (Figure~\ref{fig:noisy-two-face}, bottom). Interestingly, even though the Bayes-Optimal classifier is robust, the CNN learns a vulnerable classifier. We will return to this point in section~\ref{sec:experiments} using our realistic MFA datasets.

{}

\begin{figure}
\centerline{\small
\begin{tabular}{c}
   Male/Female Asymmetric: (${L2}=0.9$)\\
   \includegraphics[height=2.3cm]{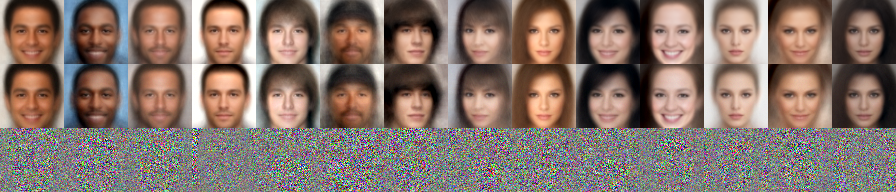}
   \hspace{2mm}
   \includegraphics[height=2.3cm]{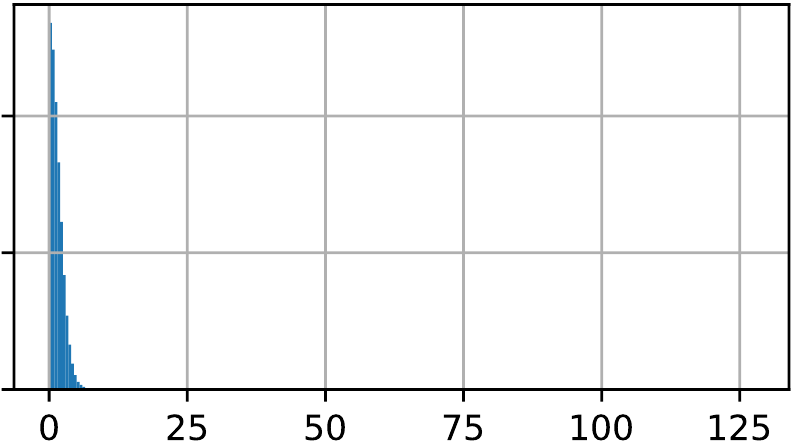}\\
     Male/Female Symmetric: (${L2}=3.3$) \\
    \includegraphics[height=2.3cm]{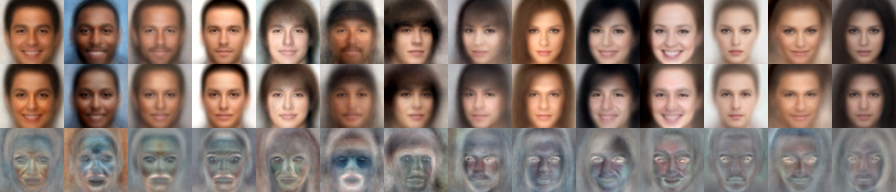} 
   \hspace{2mm}
    \includegraphics[height=2.3cm]{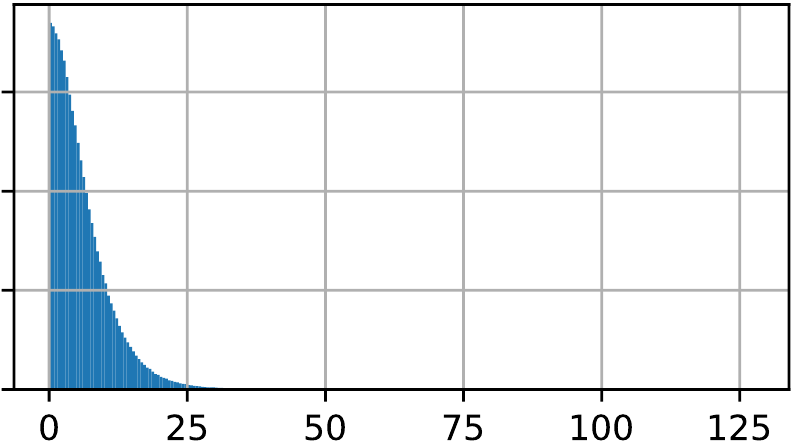}
      \end{tabular}}
\caption[]{The vulnerability of the optimal classifier depends on the presence of asymmetries. Each result shows the original images, adversarial images and perturbations (magnified for visibility as necessary). Histogram of the perturbations in pixel values are on the right.}
\label{fig:samples-mfa-with}
\end{figure}

\subsection{Symmetric and asymmetric datasets}
\label{sec:our-datasets}

To summarize our analysis, the Bayes-Optimal classifier will be \emph{provably robust} when the covariances satisfy symmetry conditions and \emph{provably vulnerable} when there are strong asymmetries.  
We therefore created symmetric and asymmetric variants of the MFA datasets.  In the \emph{symmetric} version, we regularized the MFA so that the ``off manifold" variance $\sigma$ is small and the same in all components and the distribution approximates the conditions of Theorem 2. In the \emph{asymmetric} version, we added to each MFA model one  ``outlier'' component with a diagonal covariance with much larger $\sigma$ than all other covariances and a mean that is close to the global data mean. This version approximates the conditions of Theorem 3. In the next section we attack these optimal classifiers and compare the empirical robustness with our theoretical analysis.

\subsection{Evaluating the robustness}
\label{sec:evaluation}

Unlike our theoretical results, deciding whether or not a real classifier is robust or not requires an operational definition of what constitutes a ``tiny" or ``imperceptible" perturbation. We follow the standard practice of calculating the mean perturbation $\ell_2$ norms of an adversarial attack~\citep{Schott2018a,EvaluatingCarlini}.  We allow the adversary an unlimited budget in attacking the classifiers, and measure how large a perturbation was required to cross the decision boundary. The mean is calculated only over successful attacks, when the original sample was correctly classified and the adversarial example was not. Since this definition is sensitive to outliers and the particular choice of Euclidean norm, we also examined the histograms of changes made to each pixel in the adversarial attack. Finally, we visually inspected the adversarial images.

In all 15 datasets, we found that these three methods of defining robustness are consistent. For the face images,  when the mean $\ell_2$ is less than $1.5$, then the adversarial images are almost indistinguishable from the original images, and the vast majority of the pixels in the adversarial images are within $5/255$ intensity levels from their original value. On the other hand, when the mean $\ell_2$  is around $3$, then the adversarial images are perceptually quite different from the original  ones, and many pixels differ by more than $5/255$ from their original values. 
We used a simple gradient attack in which we take small steps in the direction of the gradient of the MFA log likelihood (details in appendix~\ref{apx:attacks}). Similar results are achieved with a standard implementation \citep{papernot2018cleverhans} of the \emph{CW-L2} attack \citep{Carlini017}.

As shown in Figures~\ref{fig:samples-mfa-with} (similar to Figure~\ref{teaser-fig}b,c), the difference between the symmetric datasets and asymmetric datasets is dramatic (see appendix~\ref{apx:more-results} for similar results on other classes). When there exists a large asymmetry between the minimal variances of different Gaussians, the conditions of Theorem~3 hold, and a tiny imperceptible change is sufficient to fool the Bayes-Optimal classifier. However, when all Gaussians have the same minimal variance, the conditions of Theorem~2 hold, any adversary will need to make much larger changes and the adversarial examples become perceptually meaningful.

We now turn our attention to classifiers trained on our \emph{symmetric} datasets and ask -- will they be robust like the optimal classifiers for these datasets, or vulnerable.

\section{Experiments: Why are CNNs So Brittle?}
\label{sec:experiments}

\begin{figure}
\centerline{
\begin{tabular}{cc}
Asymmetric distributions & Non-optimal learning\\
\includegraphics[width=0.3\linewidth]{./Figs/elliptic-asymmetric.pdf} &
\includegraphics[width=0.3\linewidth]{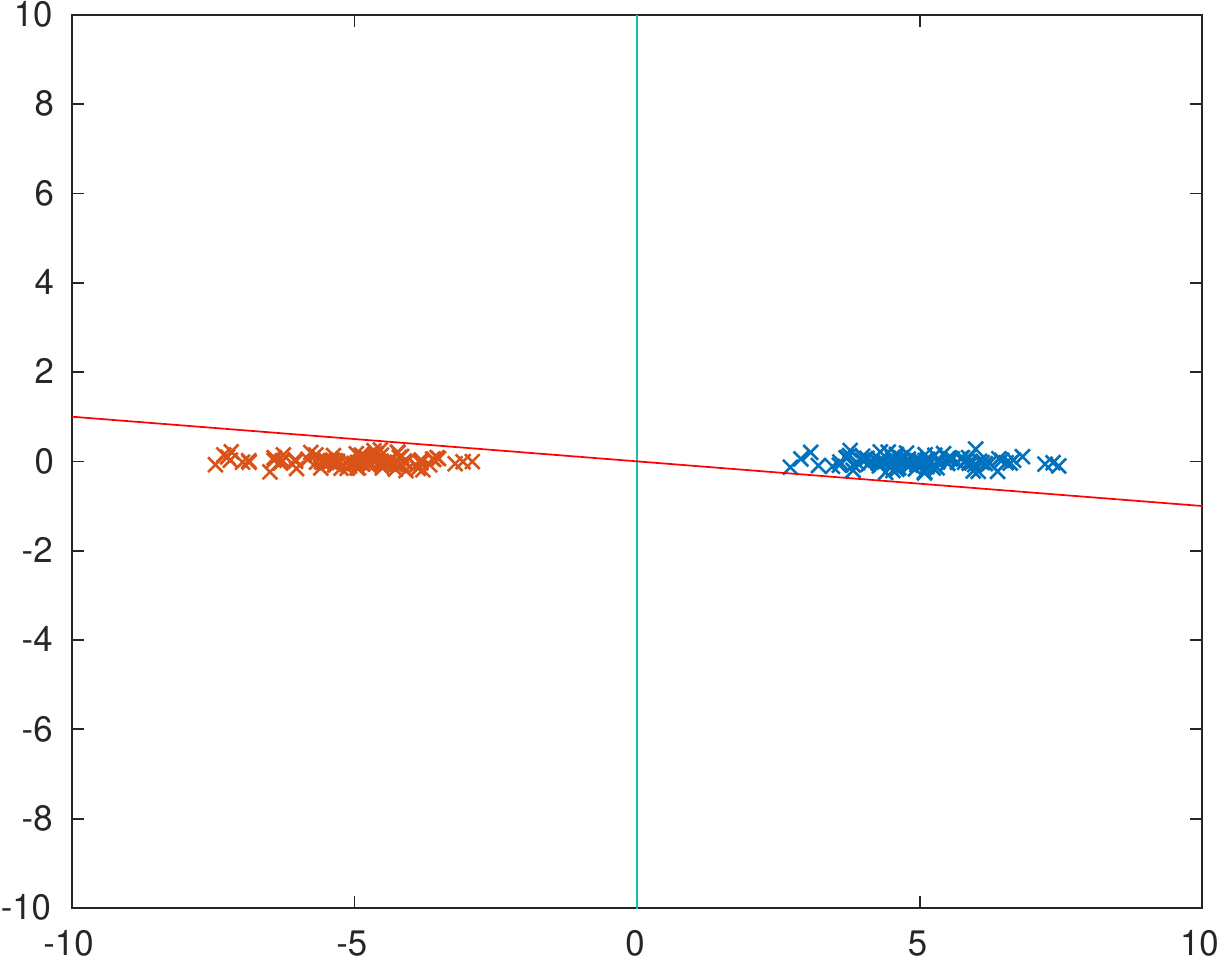} 
\end{tabular}
}
\caption[]{Why are CNNs so brittle? Is it because the data distributions are asymmetric so that the optimal classifier is also brittle (left) or is it due to non optimal learning (orange line, right) in cases where the optimal classifier is robust (blue line, right)?}
\label{fig:why-brittle}
\end{figure}

\begin{figure}
\centerline{
\begin{tabular}{cc}
CelebA & MNIST\\
\includegraphics[height=4cm]{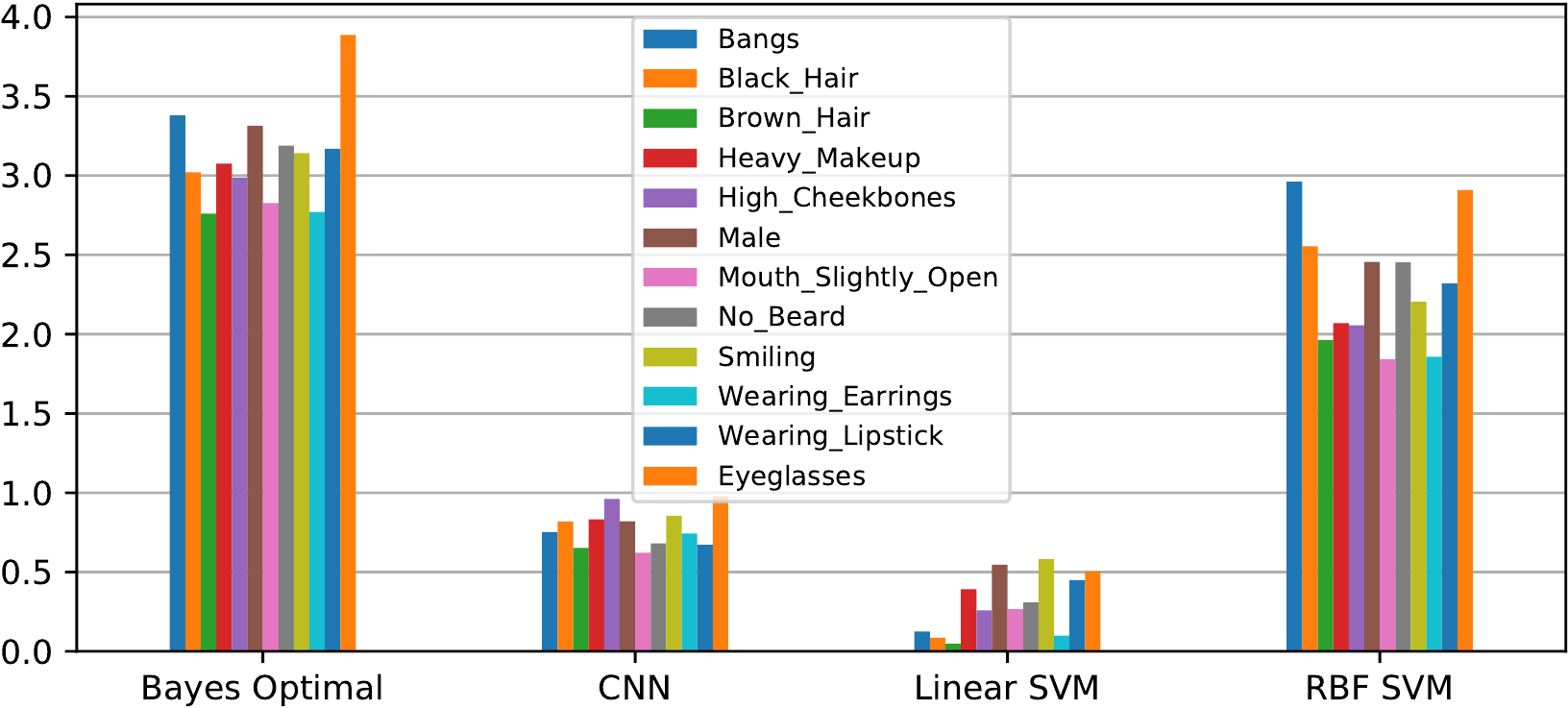} &
\includegraphics[height=4cm]{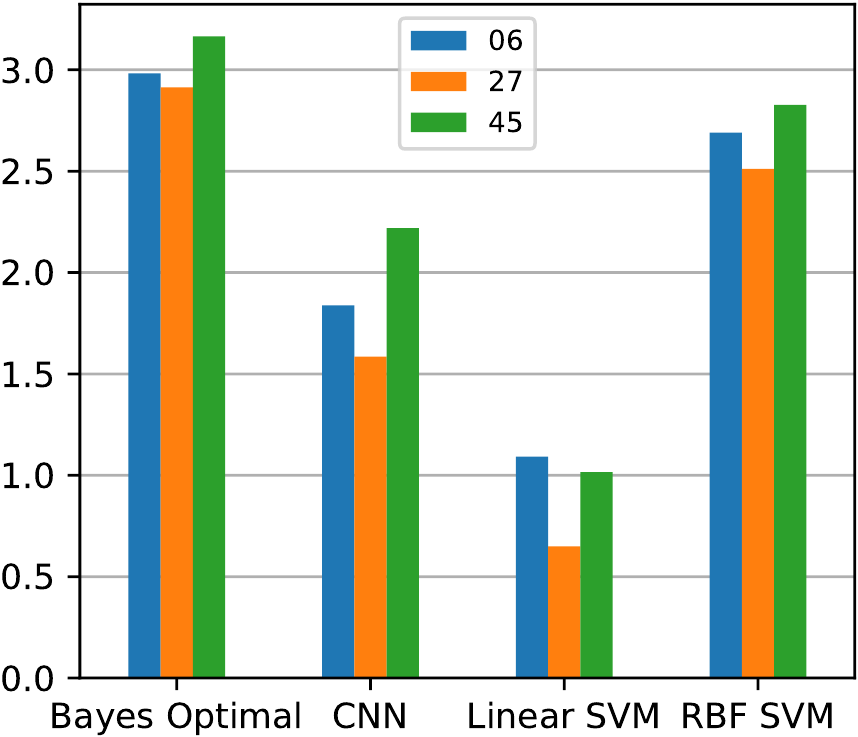} 
\end{tabular}
}
\caption[]{Adversarial attack perturbation sizes (mean L2 norm) for different models for all symmetric datasets.}
\label{fig:experiments-results}
\end{figure}

\begin{figure}
    \centering
    \begin{scriptsize}
    \begin{tabular}{rcc}
         Original: & \includegraphics[height=11mm]{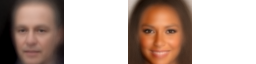} & \\ 
         Optimal: & \includegraphics[height=11mm]{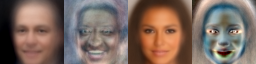} & \includegraphics[height=11mm]{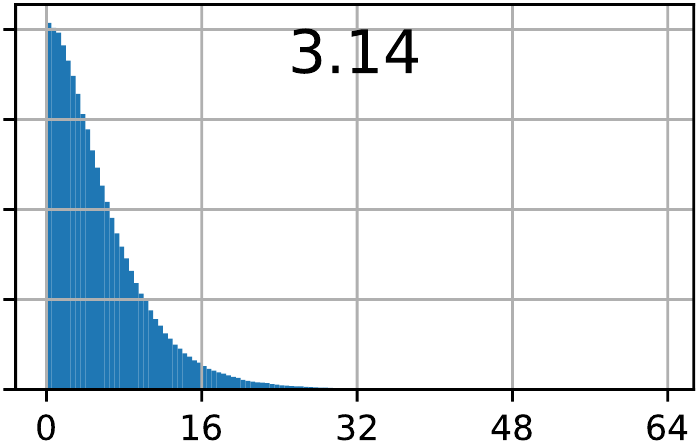}\\ 
         CNN: & \includegraphics[height=11mm]{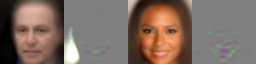} & \includegraphics[height=11mm]{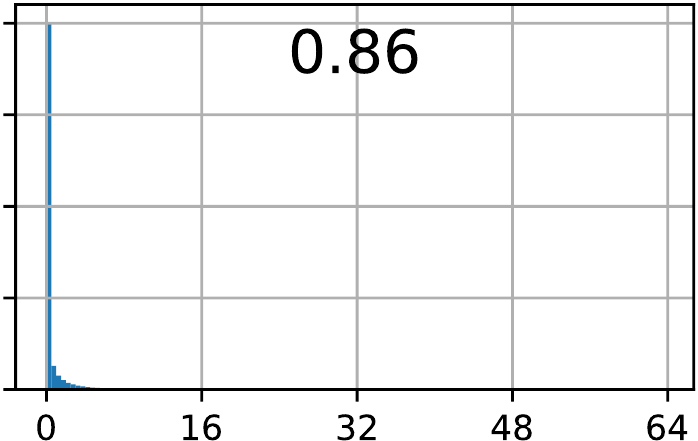}\\ 
         Lin. SVM: & \includegraphics[height=11mm]{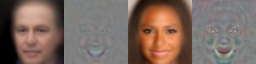} & \includegraphics[height=11mm]{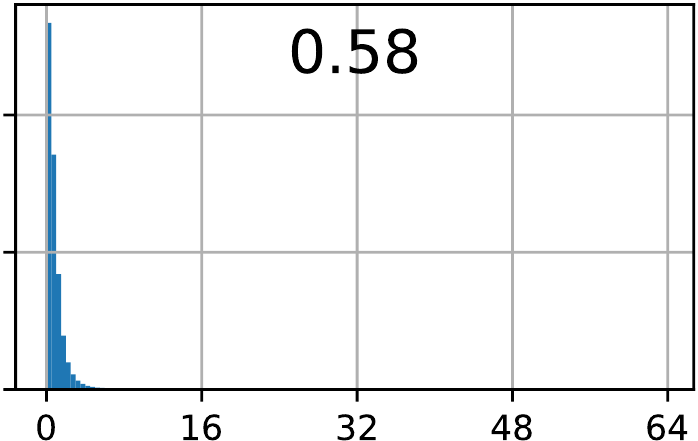}\\ 
         RBF SVM: & \includegraphics[height=11mm]{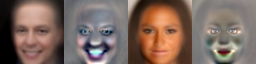} & \includegraphics[height=11mm]{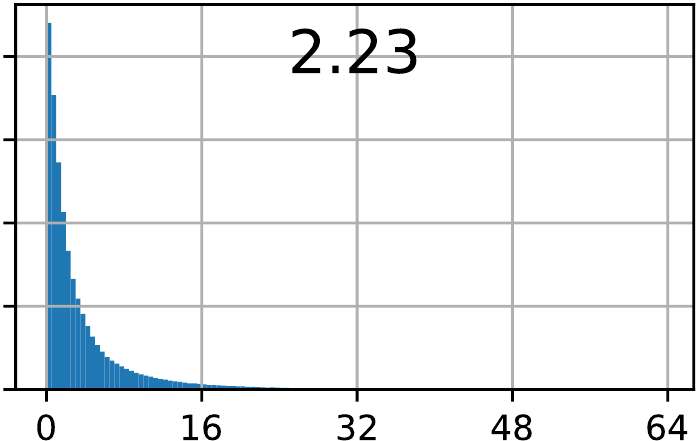}
    \end{tabular}{}
    \end{scriptsize}
    \caption{Adversarial examples, perturbations and histograms (with mean L2 value) for CelebA attribute 'Smiling'.}
    \label{fig:attacks-trained-models}
\end{figure}{}

Given our analysis, the fact that modern machine learning methods are often susceptible to tiny adversarial perturbations may be due to two very different reasons (Figure~\ref{fig:why-brittle}). One reason could be that the data distribution is asymmetric, so that the Bayes-Optimal classifier is not robust, and hence it is not surprising that a CNN is also not robust. A second possible reason is illustrated in Figure~\ref{fig:why-brittle} (right): here the data distribution is symmetric and the Bayes-Optimal classifier is robust, yet SGD starting from a bad initial condition finds a brittle classifier. If this is the case, then the brittleness is not due to the data distribution but rather a failure of the learning method. 

In order to separate the contribution of the dataset from the estimation method in the vulnerability of machine learning methods, we trained a CNN on samples from all 15 \emph{symmetric} datasets described in section~\ref{sec:our-datasets}, and measured the vulnerability of the learned CNN. We used the CNN implementation and the \emph{CW-L2} attack from the \emph{CleverHans} library \citep{papernot2018cleverhans} (details in appendices~\ref{apx:models},~\ref{apx:attacks}). We asked: will the CNN find a brittle classifier even though the optimal one is robust?

\begin{figure}
\centerline{
\begin{tabular}{ccc}
\includegraphics[width=0.33\linewidth]{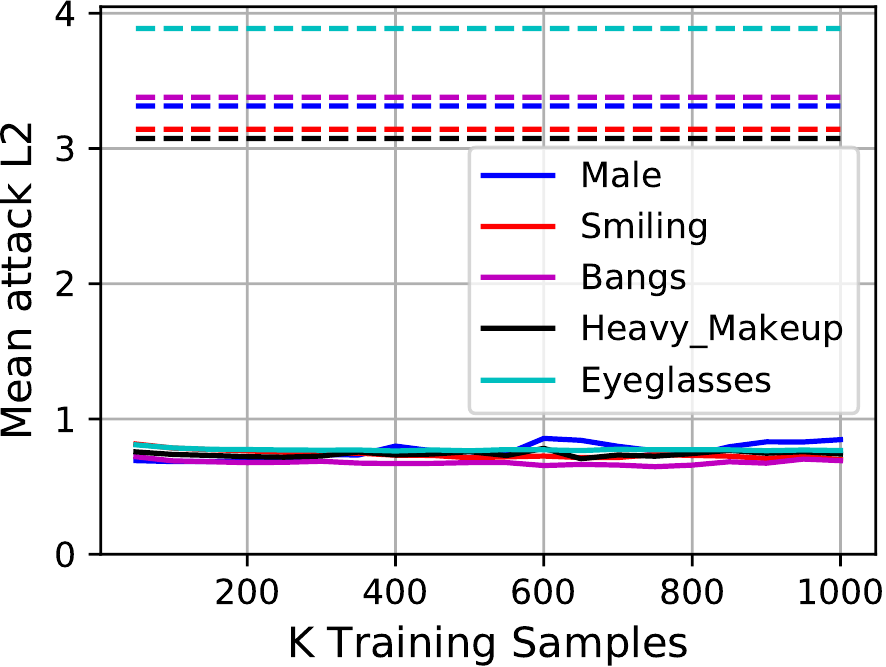} &
\includegraphics[width=0.33\linewidth]{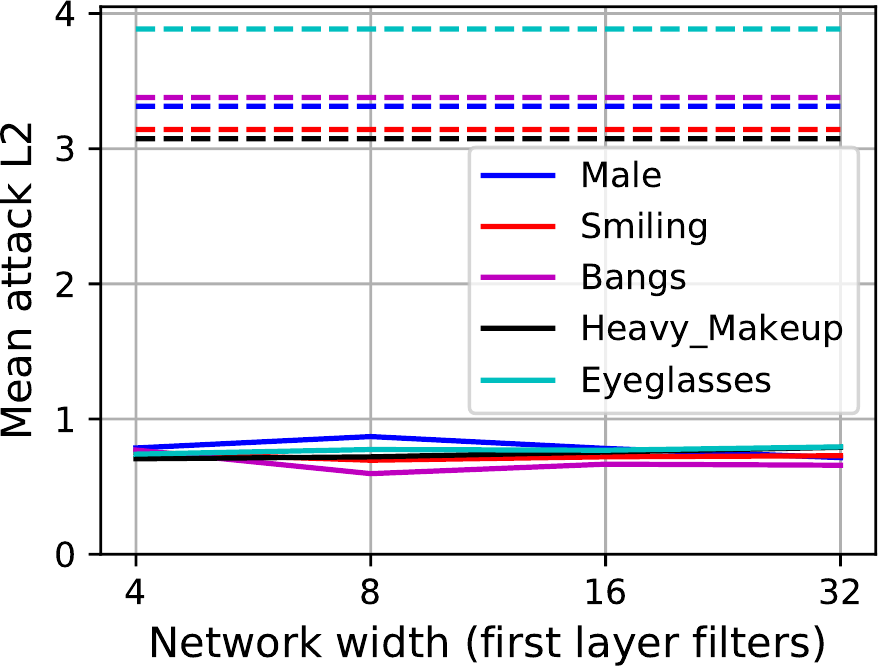} &
\includegraphics[width=0.33\linewidth]{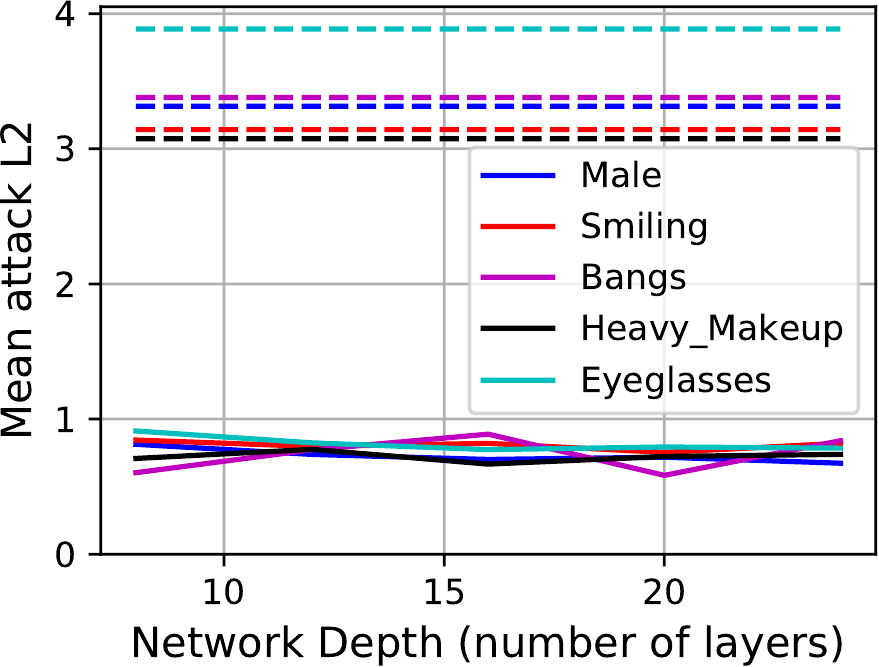}
\end{tabular}
}
\caption[]{Wider and deeper CNNs and longer training with more data does not improve the robustness (optimal classifiers are shown as dashed lines for reference).}
\label{fig:more-cnn-training}
\end{figure}

Results are shown in Figures~\ref{fig:experiments-results} and~\ref{fig:attacks-trained-models}. {\em In all 15 cases, the CNN found a high accuracy classifier that was vulnerable to small adversarial perturbations, even though the optimal classifier is robust.}  The difference is most dramatic in the CelebA tasks, where the CNN adversarial examples are almost indistinguishable from the original images (examples of the attacks on different datasets are shown in appendix~\ref{apx:more-results}). 
While there are many possible architectures and optimization methods for CNNs, we did not find any improvement in the CNN robustness in our attempts to change the number of filters, layers, training iterations etc (Figure~\ref{fig:more-cnn-training}). In particular,~\citet{Schmidt2018} have argued that one needs more training examples to achieve robust classification, so we systematically varied the amount of training images (generated dynamically at each SGD iteration), and found no significant improvement in robustness as we increased the number of training examples up to 1 million examples. 

Is it possible to learn a robust classifier for these datasets from finite training data? To answer that question, we then trained linear and RBF Support Vector Machine (SVM) classifiers on exactly the same datasets. The linear SVM attempts to maximize the margin while maintaining high accuracy, but since the optimal classifier is nonlinear it ends up learning a brittle classifier. More importantly, {\em with an appropriate bandwidth parameter RBF SVMs find robust classifiers when trained on exactly the same data} (when the bandwidth parameter is too large, the RBF performs similarly to a linear SVM).

\begin{figure}
\centerline{
\includegraphics[width=0.5\linewidth]{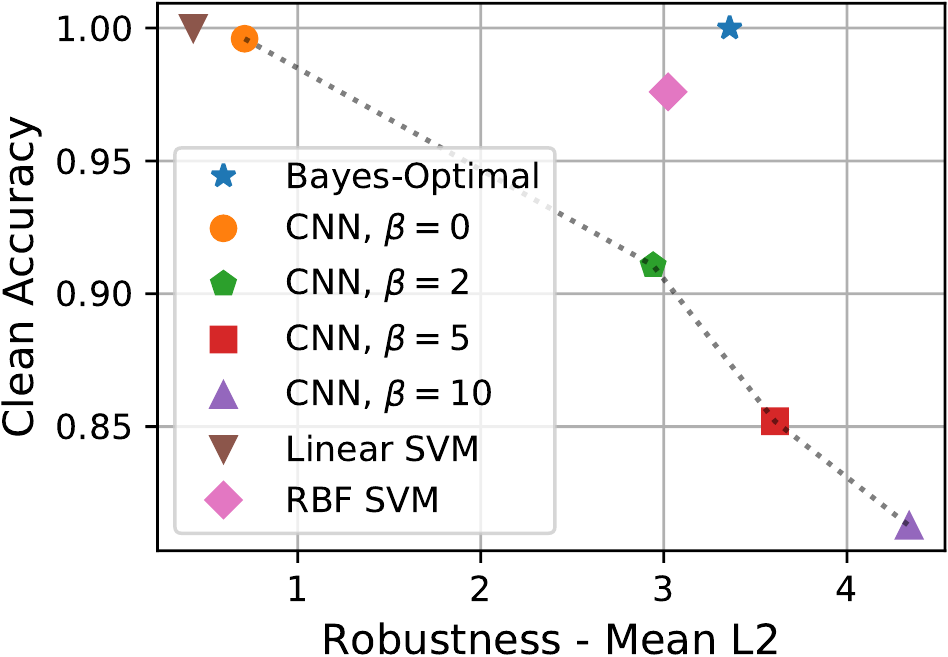}
}
\caption[]{Accuracy/robustness tradeoff (means over the different datasets) in adversarially-trained CNNs (\citealt{Zhang2019} $\beta$ values) vs.\ models that are both accurate and robust -- the Bayes-optimal and RBF SVM.}
\label{fig:accuracy-robustness}
\end{figure}

The robustness of CNNs can be improved using \emph{adversarial training}, in which adversarial samples (of some selected attack) are injected during training. In Figure~\ref{fig:accuracy-robustness} we evaluate the robustness of CNNs trained using TRADES \citep{Zhang2019} (first place in the NeurIPS 2018 adversarial challenge out of $\approx$ 2000 submissions). This algorithm has a tradeoff parameter  $\beta$ that trades off accuracy and robustness. In the only parameter setting  that yielded a robust classifier, this approach gives significantly reduced accuracy compared to the Bayes-Optimal classifier or to the RBF SVM.

Returning to Figure~\ref{fig:why-brittle}, our results strongly support the hypothesis that for these cases brittleness is due to suboptimal learning methods, even when the data distributions are symmetric and the optimal classifier is robust.

\subsection{Training and testing on real data}
One can ask, to what extent our analysis and experiments represent adversarial attacks on models trained on real data? To answer this question we trained CNNs and SVMs on five real CelebA attribute datasets, and indeed, as shown in Figure~\ref{fig:train-on-real-data} and in appendix~\ref{apx:more-results}, the results are similar to the synthetic symmetric datasets -- CNN and Linear SVM are vulnerable while RBF SVM is robust and can only be fooled when the perturbations are  perceptually meaningful.

\begin{figure}
\centerline{
\includegraphics[width=0.65\linewidth]{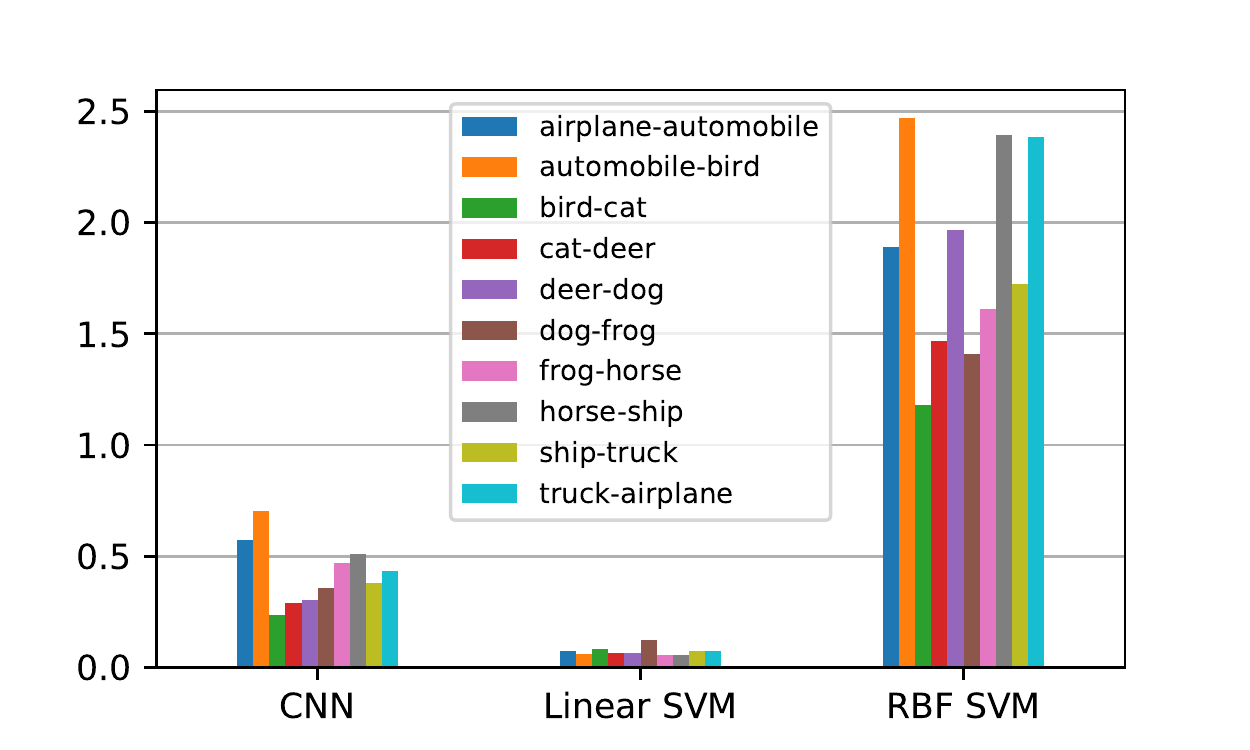}
}
\caption[]{Robustness (mean required perturbation $\ell_2$ sizes) for different binary classifiers trained on CIFAR-10 class pairs (class $i$ vs class $i+1$). Results are consistent with our other experiments.}
\label{fig:l2-cifar}
\end{figure}

To verify that our findings are not limited to specific datasets (faces, digits), we trained similar models on the more complex CIFAR-10 dataset \citep{krizhevsky2009learning}. We trained multiple binary classifiers on different class pairs and compared their adversarial robustness. As can be seen in Figure~\ref{fig:l2-cifar} the results agree with all other experiments. \footnote{Accuracy does not necessarily correlate with vulnerability: The RBF-SVM clean accuracy was on average 7\% lower then the CNN's. Linear-SVM, which was the most vulnerable, had 25\% lower accuracy compared to the RBF.}

Note that unlike our proposed symmetric data, in which the ``gold standard'' optimal classifier is both robust and accurate, on real data, which may contain variance asymmetries, different models might reach different trade-off points between accuracy and robustness. In particular, it is known that RBF SVM accuracy is highly influenced by the hyperparameters $C,\gamma$ and we performed only a minimal search to obtain these results. In future work, it would be interesting to explore the full regularization path as in~\citep{hastie2004entire}.

\begin{figure}
\centerline{
\scriptsize
\begin{tabular}{cc}
\begin{tabular}{rc}
Original: & \includegraphics[height=11mm]{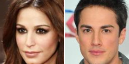}\\
CNN: & \includegraphics[height=11mm]{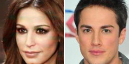} \\
Lin. SVM: & \includegraphics[height=11mm]{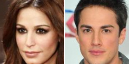} \\
RBF SVM: & \includegraphics[height=11mm]{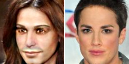} \\
\end{tabular} &
\raisebox{-0.5\totalheight}{\includegraphics[width=0.35\linewidth]{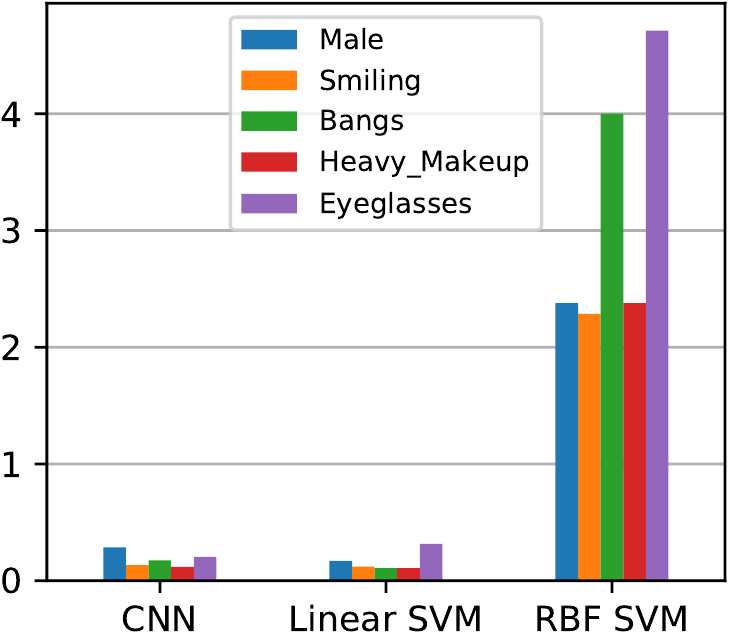}}
\end{tabular}
}
\caption[]{Adversarial examples for class Male/Female (left) and mean perturbation sizes (right) for different models trained on real CelebA images -- results are consistent with our other experiments.}
\label{fig:train-on-real-data}
\end{figure}

\section{Related Work}
\label{sec:related}
One of the first explanations of adversarial examples in CNNs was that
the decision surface learned by neural networks is “discontinuous to a
significant extent”, analogous to an attempt to discriminate the
rational numbers from the rest of the real numbers~\citep{SzegedyZSBEGF13}. However as shown by our analysis, when there are strong
asymmetries in the variances of the two classes, adversarial examples can fool
the optimal classifier even when the decision boundary is smooth and continuous. 

A second prominent theory 
suggests that the problem is that neural network classifiers are
“unreasonably linear” combined with the fact that they operate in high
dimensions~\citep{GoodfellowSS14,Goodfellow2018, gilmer2018adversarial, fawzi2018adversarial, dohmatob2018limitations, mahloujifar2019curse}. In high dimensions the
output of a random linear classifier can be changed by making a small
change in the $\ell_\infty$ norm of an example. 
\citet{FawziFF18} show a
connection between the error rate of linear and quadratic classifiers
and the adversarial vulnerability and show that linear classifiers in
high dimensions must be vulnerable for data that is not linearly
separable. \citet{ford2019adversarial} show that adversarial vulnerability is
closely related to the lack of generalization to random perturbations
and that in high dimensions even moderate failures to generalize to
high amounts of noise imply the existence of adversarial
examples. \citet{shamir} have also focused on the geometry of
high dimensions arguing that adversarial attacks may be a “natural
consequence of the geometry of $R^n$ with the $L_0$ (Hamming)
metric”.

Both our analysis and our experiments suggest that high
dimensionality is neither necessary nor sufficient for
vulnerability. When strong asymmetries exist, even two dimensional
datasets can be constructed such that the optimal classifier is
vulnerable. At the same time, when there are no asymmetries, the
optimal classifier is robust, even in very high dimensions. The same
is true for ``excessively linear'' classifiers: the RBF SVM is
probably no less linear than a CNN, yet it is robust in our symmetric and in real
datasets.

The accuracy-robustness tradefoff has also been suggested as an
explanation for adversarial vulnerability~\citep{Zhang2019}. \citet{Schmidt2018} present a
model under which adversarially robust generalization requires more
data. As our analysis shows, for symmetric datasets there is no
tradeoff between accuracy and robustness and the optimal classifier in
terms of accuracy is also robust. Our experiments also show that with
proper regularization (i.e. a RBF SVM), one can learn adversarially
robust classifiers with the same amount of data for which CNNs learn a
vulnerable classifier.

Most recently, \citet{ilyas2019adversarial}
have argued that adversarial examples are a
feature, not a bug, and showed that one can in fact obtain information
about the true decision boundary from adversarial examples. Their
analysis suggests that vulnerability results from the presence of
predictive features that are not robust. They presented a synthetic
dataset which was constructed to not contain such features, and showed
that CNN training on that dataset was  robust.  Our analysis in terms
of symmetric vs. asymmetric datasets is similar to theirs but more
general (they only considered linear classifiers). Our experimental
results, however, are quite different and suggest that in more
challenging settings, CNNs consistently learn  vulnerable classifiers
even when the asymmetries do not exist in the data distributions.

In~\citep{nakkiran2019a},   a
synthetic dataset was presented without nonrobust features, but CNN
training led to vulnerable classifiers when the dataset was
noisy. This result is consistent with \citet{tilting} 
who show that adversarial vulnerability is related to overfitting in
learning algorithms that are not sufficiently regularized. Similarly, \citet{lyu2020gradient} show theoretically and experimentally that the details of the training procedure can significantly change the
robustness of CNNs.
Our experimental results also highlight the need for regularization  in more realistic and challenging settings (including real data), while our analysis points out that lack of robustness may also occur with no overfitting when the data is asymmetric. Our experiments also show that regularization is not sufficient: the linear SVM also attempts to maximize the margin, but due to its limited expressive power it still ends up learning a vulnerable classifier.  

The analysis presented by \citet{shafahi2018adversarial} agrees with ours: that it is not merely the dimensionality of the problem but also the distribution of images that determines vulnerability. Unlike Shafahi et al., we present a realistic yet tractable model to model the distribution of images and also present experimental evidence that in many cases CNNs learn a vulnerable classifier even when the optimal classifier for that dataset is robust.

Several works analyze adversarial attacks under the \emph{manifold hypothesis} -- assuming the data resides on a low-dimensional manifold embedded in the high-dimensional image space. \cite{khoury2018geometry} claim that adversarial attacks are caused by the \emph{codimension}, the large difference between the manifold and data dimensions. Our analysis uses a more general probabilistic reasoning and shows that adversarial examples exist regardless of the codimension. \cite{zhang2020principal}  explain adversarial examples as ``perturbations perpendicular to the tangent plane of the data manifold''. They propose a PCA-based model-free attack. The paper claims that under the manifold assumption, ``it is clear that data points in adversarial regions impose a potential threat to all classifiers''. In contrast, we show that in some data distributions (even under the manifold assumption), the optimal classifier and some trained classifier are robust and only CNNs are vulnerable. \cite{jha2018detecting} detect adversarial examples by their distance from the data manifold. This work is orthogonal to ours -- we do not analyze the detectability of adversarial examples. \cite{stutz2019disentangling} generate on-manifold adversarial examples by first encoding the original image and then perturbing the latent code. In our setting, all adversarial examples are off-manifold and still we show that CNNs fail even when the optimal classifier is robust.

Intuitively, we might expect classifiers based on generative models (e.g.\ ``analysis by synthesis"~\citealt{Schott2018a}) to be more robust to adversarial attacks, since they model all the data, not just the discriminative features. But our analysis shows that even when such a classifier is based on the {\em true} generative model, it can be arbitrarily vulnerable, when the two distributions show strong asymmetries.

\section{Discussion}
Since the discovery of adversarial examples for CNNs there has been much discussion whether they are a ``bug" that is specific to neural networks or a ``feature" of high dimensional geometry. 
As far as we know, all previous works analyzed CNN vulnerability either on toy domains (e.g.\ points on a grid, concentric spheres) that are less suitable for testing CNNs, or on real image data for which a robust and accurate classifier may not exist. Unlike these works, our analysis enabled us to construct datasets that are both realistic and provably robustly-separable by the optimal classifier. This unique setting enables differentiating between unavoidable data-related vulnerability to causes related to the specific model or training method, and indeed we show that CNNs fail to reach the accurate and robust classifier in ALL our \emph{symmetric} datasets, even as we varied the parameters of the architecture and training. When applying a state-of-the-art adversarial defence method (TRADES), the achieved robustness-accuracy tradeoff is significantly inferior compared to the Bayes-Optimal and the RBF SVM classifiers.
This suggests that, at least in some situations, the presence of adversarial examples represents a failure of current, suboptimal learning methods, rather than being an unavoidable property of learning in high dimensions.

We are by no means advocating a return to using RBF SVMs. Rather, we believe that explicit  regularization methods for CNNs may enable learning robust classifiers while maintaining the power of deep architectures. Recent theoretical work on gradient descent methods suggests that they implicitly reward large margin classifiers in both shallow and deep architectures~\citep{SoudryHNGS18,poggio2017theory,lyu2020gradient} although convergence to a large margin classifier may require exponential time. 

In general, when trying to understand a complex effect, it is often useful to disentangle the different causes. The Bayes-Optimal perspective on adversarial examples identifies two possible causes: asymmetries in the datasets and suboptimal learning. Furthermore, it allows us to create tractable and realistic datasets in which one of the two causes can be clearly implicated. We are optimistic that this approach will be of great use in developing new learning algorithms that are practical and robust. 

\section*{Acknowledgments}
Supported by the Israeli Science Foundation and the Ministry of Science and Technology.

\newpage

\appendix

\section{Additional Proofs}
\label{apx:proofs}

\subsection{Proof of Theorem 1 -- \emph{Symmetric} Isotropic Gaussian}

{\bf Proof:}  For spherical and equal covariances, the Bayes-Optimal
classifier simply projects the data onto the direction $\mu_1-\mu_2$
and classifies a point based on whether that projection is closer to
the projection of $\mu_1$ or the projection of $\mu_2$. This means that the problem reduces to a scalar problem, with two distributions whose means are at distance $\|d\|$ and have scalar variance $\sigma^2$. With high probability, under the assumption, all the points are close to one of the means, so they are at distance $\|d\|/2$ from the decision boundary. $\eop$

\subsection{Proof of Theorem 2 -- \emph{Symmetric} MFA}

We will first handle the simpler case where each class is a single Factor Analyzer:

{\bf Lemma A.1: \emph{Symmetric} low-rank $+$ diagonal.}  Suppose both classes have a 
covariance that is a sum of a low-rank matrix plus a diagonal matrix
$\Sigma_i=A_i A_i^T + \sigma^2 I$ and $A_1,A_2$ have the same singular values and $\sigma$ is the same for both classes.  Let $\bm{d}$ be the minimal distance between the two linear subspaces:
\[
\bm{d}= \min_{z_1,z_2} \| \mu_1 + A_1 z_1 - (\mu_2 + A_2 z_2)\|
\]
As $\sigma \rightarrow 0$ then almost all points are at distance $\bm{d}/2$
from the Bayes-Optimal decision boundary.

{\bf Proof:} If $x \sim N(\mu,AA^T + \sigma^2 I)$ then the distribution of $x$ can be described by the following generative model:
\BEA
z & \sim & N(0,I) \\
\eta & \sim & N(0, \sigma^2 I) \\
x &=& Az + \mu + \eta
\EEA

We can also write the likelihood of $x$ as:
\BE
P(x) = \int_z P(x,z) dz
\EE

Since $x,z$ are jointly Gaussian, $P(x,z)$ is an unnormalized Gaussian function of $x$ and the integral is given by the height of the unnormalized Gaussian divided by the square root of the determinant of the second derivative of $\log P(x,z)$ with respect to $z$ (see for example \citealt{mackay2003information} p. 341). So we can write:
\[
\log P(x) =- \min_z \left(\frac{1}{2 \sigma^2} \|\mu + A z - x\|^2 -
\frac{1}{2} z^T z \right) - \frac{1}{2} \log \det (I + \frac{1}{\sigma^2}  A^T A) + \frac{N}{2} \log (2 \pi)
\]

Since $A_1,A_2$ are assumed to have the same singular values, the log determinant term in the log likelihood is the same for both classes so the Bayes-Optimal classifier will classify a point as belonging to class $1$ if:
\[
- \min_z \left(\frac{1}{2 \sigma^2} \|\mu_1 + A_1 z - x\|^2 -
\frac{1}{2} z^T z \right)
>
- \min_z \left(\frac{1}{2 \sigma^2} \|\mu_2 + A_2 z - x\|^2 -
\frac{1}{2} z^T z \right)
\]
As $\sigma \rightarrow 0$ the Bayes-Optimal decision rule will simply be:
\[\min_z  \|\mu_1 + A_1 z - x\|^2 
<
\min_z \|\mu_2 + A_2 z - x\|^2 
\]
so that a point is classified based on which subspace it is closer to.

Now, as $\sigma \rightarrow 0$ almost any point $x$ from class $1$ will
have distance close to $0$ to the first linear subspace and distance of at least
$\bm{d}$ to the second subspace. If we perturb $x$ by a vector $\delta$
then the distance to each subspace can change by no more than
$\|\delta\|$. This means that if $\|\delta\|<\bm{d}/2$, the distance to
the first subspace will remain smaller than the distance to the second
subspace, and hence the Bayes-Optimal classifier will not be fooled by
any perturbation whose norm is less than $\bm{d}/2$. $\eop$

{\bf Mixture Model.} We now assume that the distribution in each class can be represented as a Gaussian Mixture Model and denote by $p_{ik}$ the $k$th Gaussian in class $i$ and by $\pi_{ik}$ its prior probability. 
 We also assume that within each class, the components are well separated, i.e. that for each datapoint the assignment probabilities put all their mass on one of the components.  More formally, denote by $g_{ik}(x) = \frac{\pi_{ik} p_{ik}(x)}{\sum_j \pi_{ij} p_{ij}(x)} $ the assignment probability  of a datapoint $x$ to component $k$,
then we assume that for each $x$, $g_{ik}(x)=\delta(k-k_i^*(x))$ where $k_i^*(x)$ is the index of the component that is most likely to have generated $x$ under probability $p_i(x)$.  Under this assumption, the probability of generating a point $x$ under $p_1$ is simply $p_1(x)=\pi_{k^*_{1}(x)} p_{1 k^*_1(x)} (x)$.
We will also assume that within each distribution, the assigned component does not change when we perturb $x$ by a perturbation $\delta$ smaller than $d/2$: $k_i^*(x)=k_i^*(x+ \delta)$.

{\bf Proof:}  By the well separateness assumption, the optimal classifier simply compares the likelihood of a point $x$ under the most likely Gaussian component in each class. This means we can directly apply Lemma~A.1, where $p_1,p_2$ are the most likely Gaussian component in each class. $\eop$

\subsection{Proof of Theorem~3 -- \emph{Asymmetric} GMM}

We will first prove Lemma 1 -- \emph{Asymmetric} Gaussian:

{\bf Proof:} We denote by $\sigma_2 = d^T \Sigma_2 d$ the variance of the data in direction $d$ under the distribution of the second class $p_2$. Note that by the assumption that $\Sigma_2$ is full rank, this variance must be nonzero. We write the vector $x$ as $(t,s)$ where $t$ is the
projection in direction $d$ and $s$ is a vector of projections in directions
orthogonal to $d$. We denote by $t_i$ the projection of $\mu_i$ in direction $d$.  The decision surface as a function of $t$ is a
solution to:
\BEA
\label{eq:decision-gaussian}
\left( \frac{1}{\sigma_1^2} - \frac{1}{\sigma_2^2} \right) t^2 +
2\left(\frac{t_2}{\sigma_2^2} - \frac {t_1}{\sigma_1^2}
\right) t + \left( \frac {t_1^2}{\sigma_1^2}- \frac
  {t_2^2}{\sigma_2^2} + \log \frac{\sigma_1^2}{\sigma_2^2} \right) = 
  \log p_2(s|t) - \log p_1 (s|t) \nonumber
\EEA
  Now since $d$ is a direction of minimal variance, it must be an  eigenvector of $\Sigma_1$ so that $s$ and $t$ are independent under $p_1$ and we can write $\log p_1(s|t)= \log p_1(s)$. Using the standard equation for conditional Gaussians, $p_2(s|t)$ will also be a Gaussian with the following mean and covariance:
  \BEA
  \mu_{s|t} &=& \mu_s + \Sigma_2^{st} (t - t_2) \\
  \Sigma_{s|t} &=& \Sigma_2^{ss} - \Sigma_2^{st} \frac{1}{\sigma_2^2} \Sigma_2^{ts}
  \EEA
  where $\Sigma_2^{ss},\Sigma_{2}^{st}$ are the appropriate submatrices of the covariance matrix $\Sigma_2$.
  As $\sigma_1 \rightarrow 0$ then $t \rightarrow t_1$ and $\mu_{s|t}, \, \Sigma_{s|t}$ will not depend on $t$. This means that the right-hand side of equation~\ref{eq:decision-gaussian} depends only on $s$ and not on $t$ or $\sigma_1$.

  As $\frac{\sigma_1}{\sigma_2}$ approaches zero, the  solutions of this
equation approach $t_1$. And because $\sigma_1 \rightarrow 0$, almost
all samples from the first Gaussian will be close to $t_1$, so moving
$x$  by a tiny amount in direction $d$ will change the optimal decision. \eop

The proof of the GMM case is now similar to that of the mixture model in Theorem 2, but using Lemma~1 instead of Lemma~A.1.

\subsection{Robustness of the Optimal Classifier -- Non Gaussian Distributions}

A natural question following Theorems~1-3 is to what extent the result depends on the Gaussian distribution. To address this, we now consider {\em discrete} distributions. We assume that every instance $x$ is described by quantized features that can take on a discrete number of values. For example, the features can be wavelet coefficients of an image that are discretized into $256$ possible values. This means that $p_1(f),p_2(f)$ are simply very large tables that give the probability of observing a particular discrete set of image features given each of the classes. Of course learning such a large table is infeasible without additional assumptions, but recall that we are analyzing the Bayes-Optimal case, where we assume $p_1(f),p_2(f)$ are known. 

{\bf Lemma A.2:} Assume there exists a feature $i$ and a quantization level $k$  so that $p_1(f_i=k) \rightarrow 1$.  Assume also that for any feature vector $f$, $p_2(f)>0$. Then almost any point in class $1$ is one quantization level away from the optimal decision boundary. 

{\bf Proof:} We again write $f=(s,t)$ where $s$ is the $ith$ feature and $t$ are all other features.
\BE
p_1(s,t)=p_1(s)p_1(t|s);
\EE
Now since $p_1(s)$ approaches $1$ for $s=k$ and $0$ otherwise, for almost any point in class $1$ the value of that feature is equal to $k$. We now change the feature by one quantization level and obtain a new feature vector $(\tilde{s},t)$ and $p_1(\tilde{s},t)=p_1(\tilde{s})p_1(t|\tilde{s}) \rightarrow 0$. On the other hand, by the assumption $p_2(\tilde{s},t)>0$, so that this point would now be classified as belonging to class 2. \eop

As in the Gaussian case, we do not need $p_1(s)$ to be exactly equal to a delta function for the decision surface to be close to most points. It is enough that $p_1(\tilde{s})$  be much smaller than the minimal values of $p_2$ for the decision to be flipped when we replace $s$ with $\tilde{s}$. Figures~\ref{figure-toy}g,c illustrate this dependence. In both cases, the data is sampled from a discrete distribution where the features are simply discretization of the two spatial coordinates into $100$ levels each. In other words, $p_1$ and $p_2$ are tables of size $10,000$ and each entry in the table represents the probability of generating a point at one of the $10,000$ possible locations.  In the top example, the minimal value of the probability table is approximately the same in both classes, while in the bottom example, there is a strong asymmetry and the decision boundary becomes close to all points in one of the classes. 

It is easy to see that Theorems~2,3 that discusses mixture distributions are applicable to the discrete case as well.

\section{Models}
\label{apx:models}

In this section we provide additional information about the different classification models -- architecture, hyper-parameters and training procedure.

\subsection{MFA}
\label{sec:mfa}
A Mixture of Factor Analyzers (MFA)~\citep{ghahramani1996algorithm} is a Gaussian Mixture Model where each component is a Factor Analyzer parameterized by a \emph{low rank plus diagonal} covariance matrix. MFA provides a good tradeoff between the non-expressive diagonal-covariance model and a full-covariance model, which is too computationally expensive for high-dimensional data such as full images. 

The model for a single Factor Analyzer component is:
\begin{align}
x = Az + \mu + \eta \text{ , }
z \sim \mathcal{N}(0,\,I) \text{ , } \eta \sim \mathcal{N}(0,\,D) \,,
\end{align}
where $A$ is the rectangular \emph{factor loading} matrix, $z$ is a low-dimensional latent factors vector, $\mu$ is the mean and $\eta$ is the added noise with a diagonal covariance $D$ (which may be isotropic: $D=\sigma^2 I)$
This results in the Gaussian distribution $x \sim \mathcal{N}(\mu,\,AA^T+D)$. The MFA is a mixture of such Gaussians. 

The MFA model was trained using the code provided by \citet{RichardsonW18}. The models are trained using Stochastic Gradient Descent. The training data (CelebA, MNIST) is first split by the desired binary attribute (e.g.\ Smiling / Not Smiling) and then a separate MFA model was trained independently for each subset of training samples. Because of imbalance in the number of samples per class in CelebA, we set the number of components as the number of samples divided by 1000. For MNIST we used a fixed value of 25 components per class. We chose an MFA latent dimension of 10 for CelebA and 6 for MNIST.

To allow attacking the MFA model with standard adversarial attacks such as CW-L2, we implemented it in TensorFlow as a standard CleverHans \citep{papernot2018cleverhans} model.

\subsection{Bayes-Optimal}
The MFA model is the Bayes-Optimal classifier when the data is sampled from that model.
We modified the MFA models that were trained for the different classes to define pairs of Bayes-optimal models -- \emph{symmetric} and \emph{asymmetric}.

\begin{figure}
    \centering
    \subfigure[]{
    \includegraphics[height=3cm]{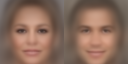}}
    \subfigure[]{
    \includegraphics[height=3cm]{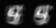}}
    \caption{Outlier components means for CelebA (a) and for MNIST (b). Samples from these components are the shown means plus strong isotropic noise with $\sigma=0.5$.}
    \label{fig:outliers}
\end{figure}

For the \emph{symmetric} models, we simply fixed all noise variance values in $D$ to a small value ($\sigma = 0.01$).
To construct the \emph{asymmetric} models we added two outlier components (one for each class) that are equal to the dataset global mean plus changes along a direction of low-variance: We performed PCA over the entire dataset and took the eigenvector for the 50th largest eigenvalue as this direction, where the mean of one outlier is in the positive direction and the mean of the other in the negative one. We set $A=0$ and $\sigma=0.5$ for both outliers, making them spherical Gaussians around the two means with relatively large noise compared to the other components. See Figure~\ref{fig:outliers} for the outlier component means for CelebA and for MNIST.

\subsection{CNN}
We used the reference CNN implementation from the CleverHans library \citep{papernot2018cleverhans}, which is a benchmark library for evaluating adversarial attacks and defences. The network consists of 2D convolution layers with a kernel size of 3 and \emph{Leaky ReLU} activations. We used a stride of 2 in several equally-spaced layers along the depth of the network to reduce the spatial dimension and at each such layer we doubled the width (number of channels). The network ends with a single fully-connected layer. All other hyper-parameters were left at their default values and the optimization method was \emph{Adam}. Our baseline small CNN achieves 100\% train and test accuracy on the symmetric datasets and we also experimented with increasing both the depth and the width of the CNN by a large factor (see Figure~\ref{fig:more-cnn-training}).

\subsection{Linear SVM}

We used the standard 2-class linear SVC implementation provided by sklearn/libsvm \citep{pedregosa2011scikit}. Linear SVM is trained directly on the vectorized image samples. The learned model consists of a weight vector $W$ and a scalar bias $b$. The decision for a sample $x$ is simply $\operatorname{sign}(W^Tx + b)$.

\subsection{RBF SVM}
We used sklearn for the Radial Basis Function (RBF) kernel SVM as well. Selection of two hyper-parameters is required, the radial kernel coefficient $\gamma = \frac{1}{2 \sigma^2}$ and $C$, a regularization term. We used  the default $C=1.0$ and the highest $\gamma$ value that still provided a high classification accuracy.

\section{Attacks}
\label{apx:attacks}

\subsection{CW-L2}

The Carlini \& Wagner L2 attack \citep{Carlini017} is a recommended strong attack that minimizes the perturbation L2 norm. The attack minimizes a weighted combination of a classification loss with the perturbation L2 size. The relative weight is a parameter that is found using a binary-search. We used the CleverHans implementation with the following hyper-parameters: 500 iterations, 3 binary-searches and a learning rate of 0.01.

\subsection{Gradient Descent Attack}

Since the MFA and SVM models provides a simple closed-form expression for the likelihood and its gradient, we implemented a simple and fast version of a gradient-attack for these models. Our attack performs multiple fixed-size steps in the direction of the gradient of the difference in log-likelihood between the source and target Gaussian components, until the decision boundary is crossed. We repeated some of the experiments with the (much slower) CW-L2 attack and verified that the results are similar (i.e. models that are shown to be robust to our gradient descent attack are also robust to the CW-L2 attack with similar perturbation magnitudes).

\section{Additional Results}
\label{apx:more-results}

In this section we provide additional experimental results for the different datasets, attributes and models.

\subsection{Distance to the Decision Surface -- Symmetric Datasets}

According to Theorem~2, if the data can be represented as a mixture of low-rank plus diagonal Gaussians and the off-manifold (diagonal) variances are all small and similar (no strong asymmetries), then the distance from a sample to the optimal decision surface will be half the distance to the nearest component subspace in the other (target) class.

\begin{figure}[ht]
\centering
\includegraphics[width=0.5\textwidth]{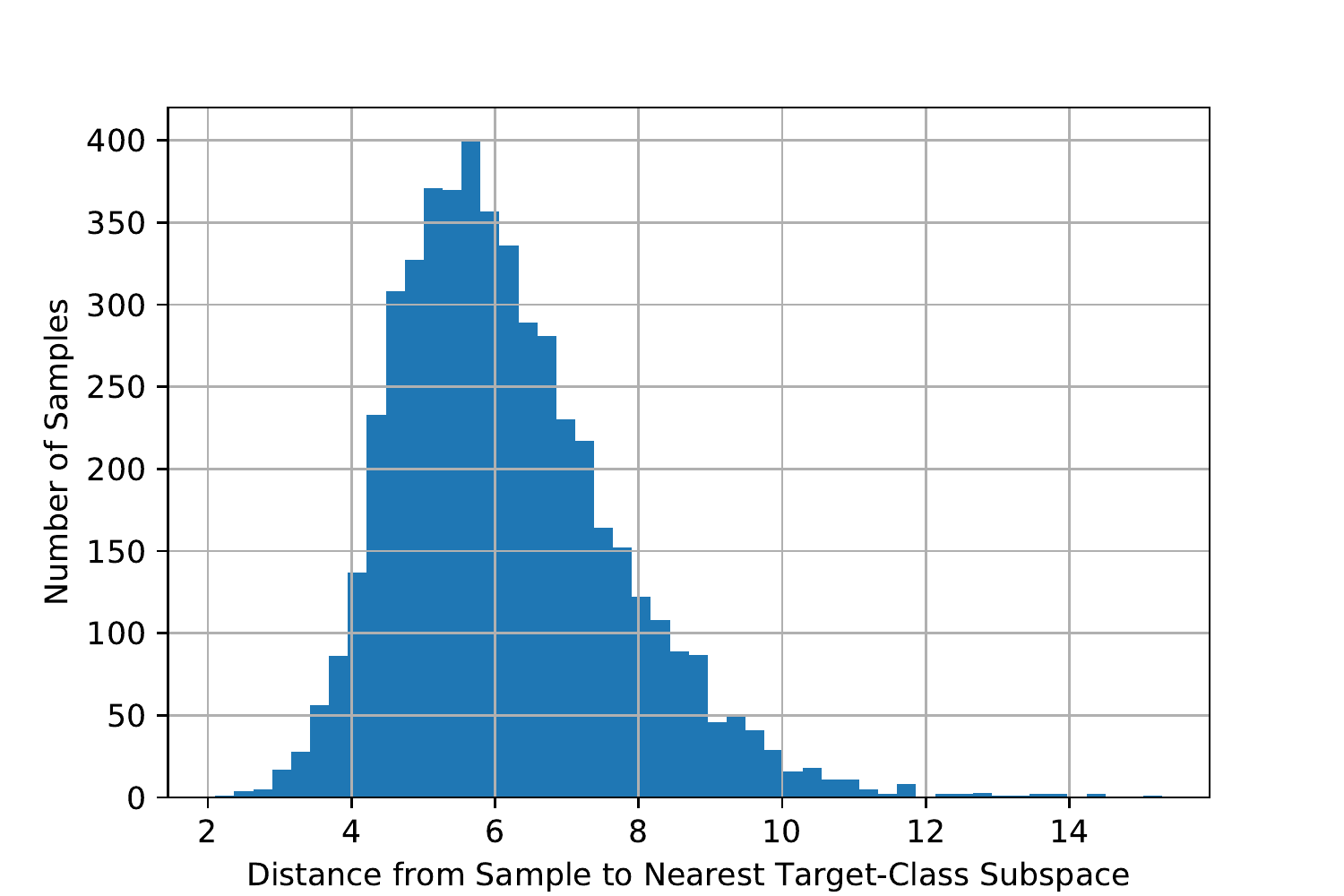}
\caption{Histogram of distances ($\ell_2$) from samples from the Male/Female \emph{symmetric} dataset to the nearest Gaussian subspace in the other class. Results are consistent with the required adversarial perturbation magnitude.}
\label{fig:distances}
\end{figure}

In toy distributions (e.g.\ Figure~\ref{figure-toy}) we can control these distances arbitrarily, but what will the distances between subspaces be in our \emph{symmetric} datasets, which approximate the manifold of real image datasets? 
As can be seen in Figure~\ref{fig:distances} (for Male/Female dataset), the distances to the nearest subspace in the other class are large -- mean of $6.2$). These values are consistent with the mean adversarial perturbation sizes that were actually required to fool the Bayes-Optimal classifier (mean $\ell_2$ of 3 -- half of the mean distance to the nearest component subspace). The same is true for the other symmetric datasets.

\subsection{Symmetric vs. Asymmetric}
\begin{figure}[ht]
\centerline{\small
\begin{tabular}{c}
      Smiling/NotSmiling Asymmetric: (${L2}=0.8$) \\
      \includegraphics[width=0.99\columnwidth]{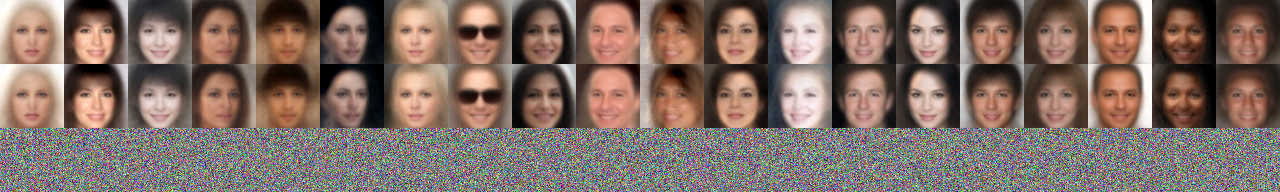} \\
      Smiling/NotSmiling Symmetric: (${L2}=3.1$)\\
      \includegraphics[width=0.99\columnwidth]{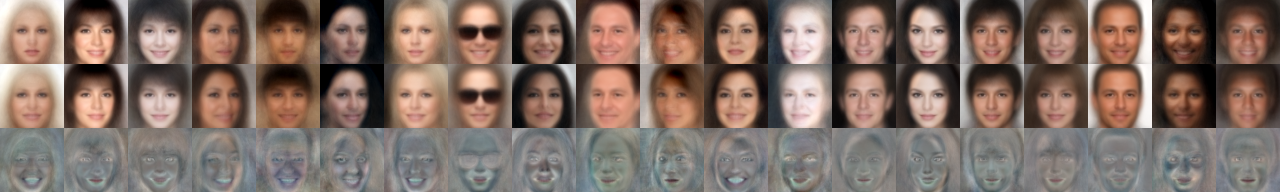} \\
      0/6 Asymmetric: (${L2}=0.5$) \\
      \includegraphics[width=0.99\columnwidth]{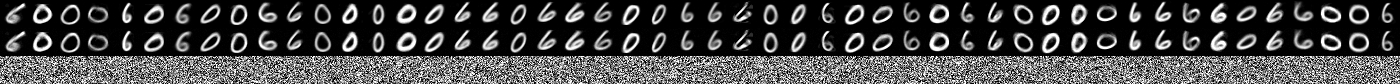} \\
      0/6 Symmetric: (${L2}=3.0$) \\
      \includegraphics[width=0.99\columnwidth]{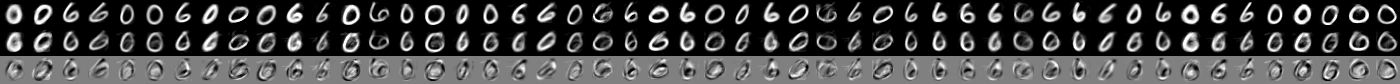} 
      \end{tabular}}
\caption[]{The vulnerability of the optimal classifier depends on the presence of asymmetries. Each result shows the original images, adversarial images and perturbations (magnified for visibility as necessary). In the symmetric datasets, attacking the optimal classifier requires large and perceptually meaningful perturbations.}
\label{fig:appendix-samples-mfa-with}
\end{figure}

Figure~\ref{fig:appendix-samples-mfa-with} presents additional examples comparing symmetric and asymmetric datasets and the relative robustness of their Bayes-Optimal classifiers to adversarial examples.

\subsection{Symmetric Datasets}

Table~\ref{tab:model-accuracy} lists the clean and adversarial classification accuracy of all models for all \emph{symmetric} datasets.

\begin{figure}
    \centering
    \begin{small}
    \begin{tabular}{rcc}
         Orig & \includegraphics[height=12mm]{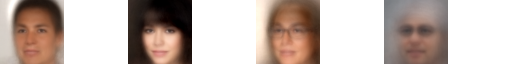} & \\ 
         Optimal & \includegraphics[height=12mm]{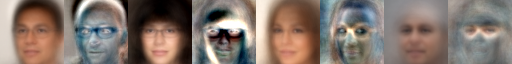} & \includegraphics[height=12mm]{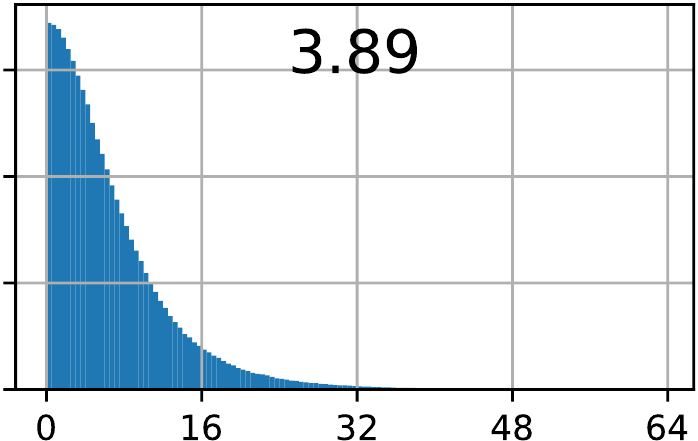}\\ 
         CNN & \includegraphics[height=12mm]{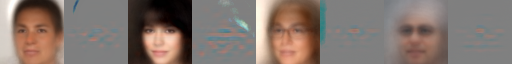} & \includegraphics[height=12mm]{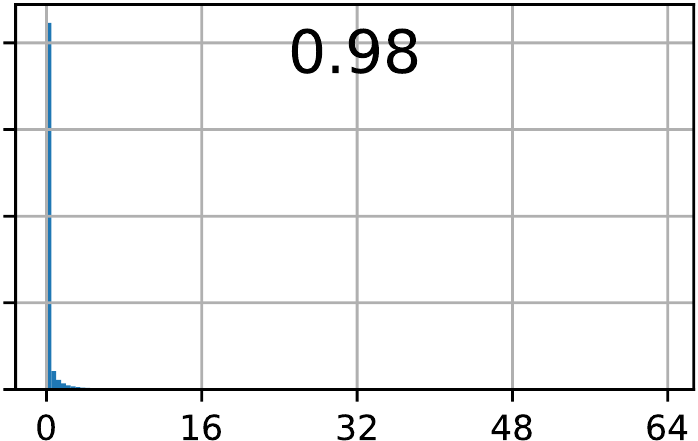}\\ 
         Lin. SVM & \includegraphics[height=12mm]{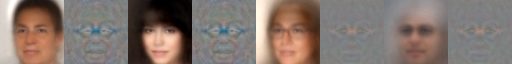} & \includegraphics[height=12mm]{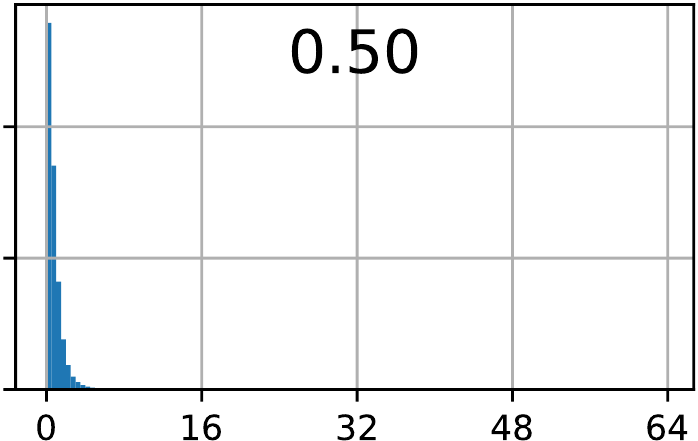}\\ 
         RBF SVM & \includegraphics[height=12mm]{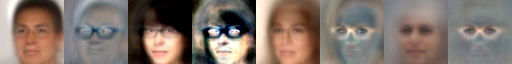} & \includegraphics[height=12mm]{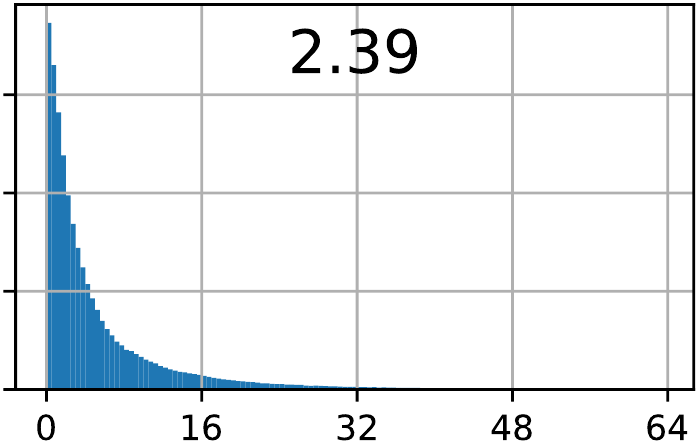}
    \end{tabular}{}
    \end{small}
    \caption{Samples, perturbations and histograms for CelebA attribute 'Eyeglasses'}
    \label{fig:res-eyeglasses}
\end{figure}

\begin{figure}
    \centering
    \begin{small}
    \begin{tabular}{rcc}
         Orig & \includegraphics[height=12mm]{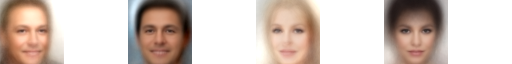} & \\ 
         Optimal & \includegraphics[height=12mm]{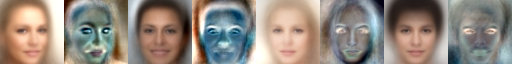} & \includegraphics[height=12mm]{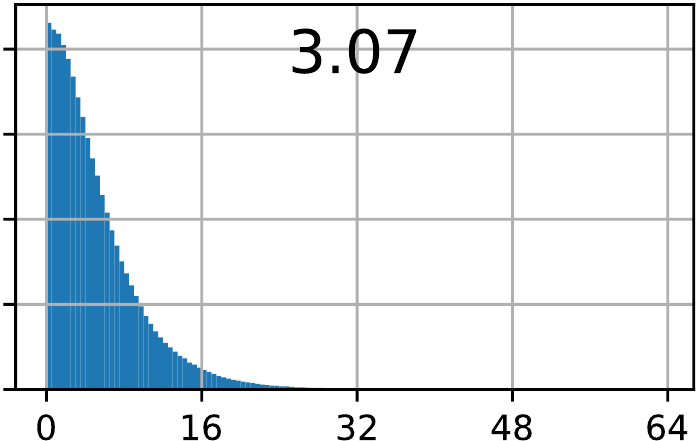}\\ 
         CNN & \includegraphics[height=12mm]{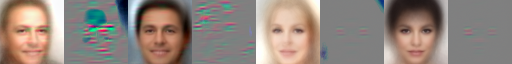} & \includegraphics[height=12mm]{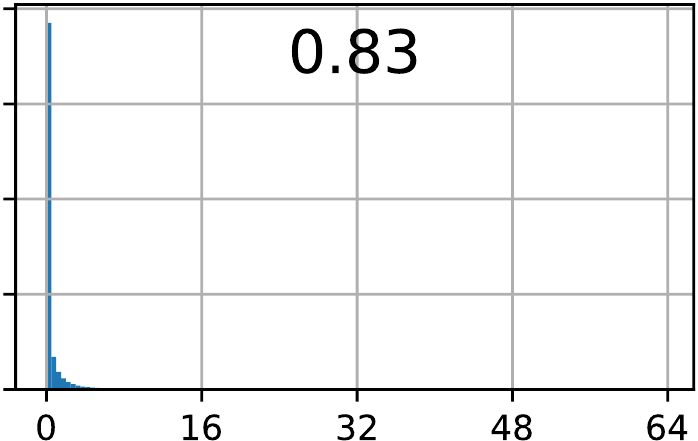}\\ 
         Lin. SVM & \includegraphics[height=12mm]{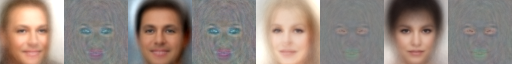} & \includegraphics[height=12mm]{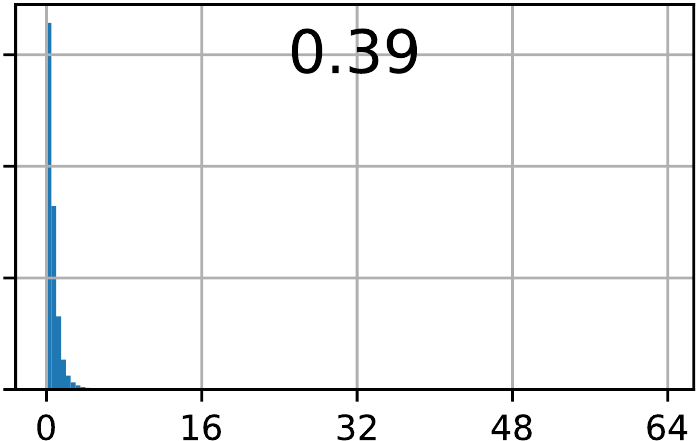}\\ 
         RBF SVM & \includegraphics[height=12mm]{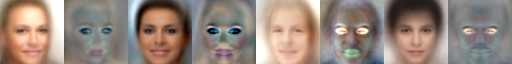} & \includegraphics[height=12mm]{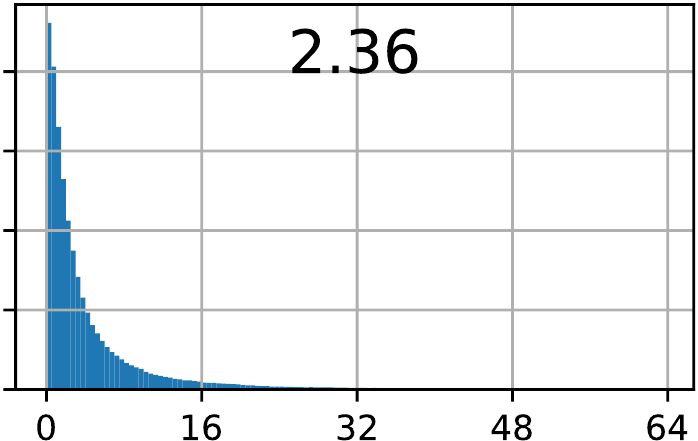}
    \end{tabular}{}
    \end{small}
    \caption{Samples, perturbations and histograms for CelebA attribute 'Heavy Makeup'}
    \label{fig:makeup}
\end{figure}

\begin{figure}
    \centering
    \begin{small}
    \begin{tabular}{rcc}
         Orig & \includegraphics[height=12mm]{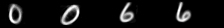} & \\ 
         Optimal & \includegraphics[height=12mm]{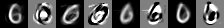} & \includegraphics[height=12mm]{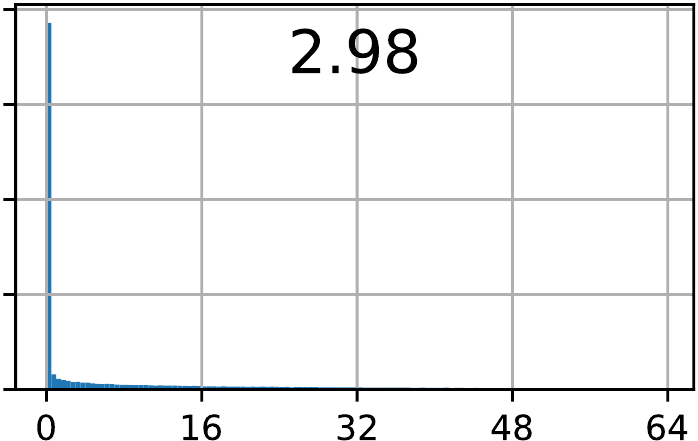}\\ 
         CNN & \includegraphics[height=12mm]{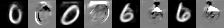} & \includegraphics[height=12mm]{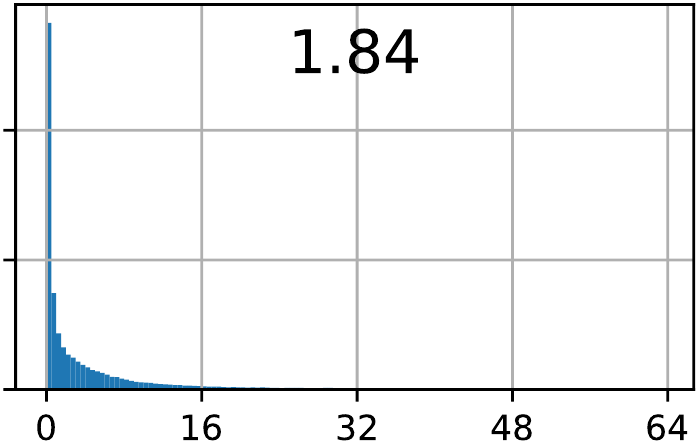}\\ 
         Lin. SVM & \includegraphics[height=12mm]{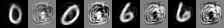} & \includegraphics[height=12mm]{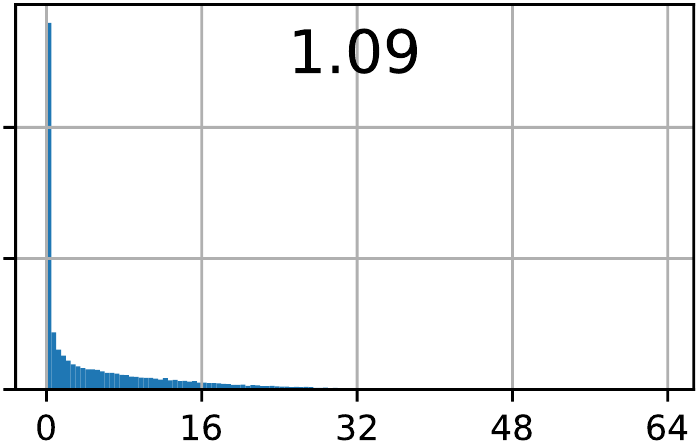}\\ 
         RBF SVM & \includegraphics[height=12mm]{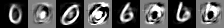} & \includegraphics[height=12mm]{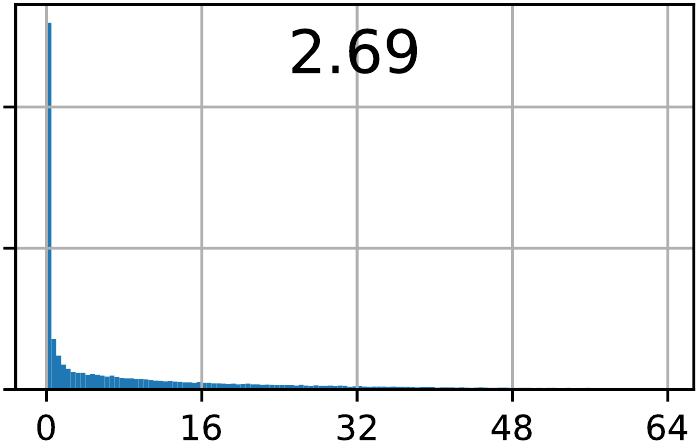}
    \end{tabular}{}
    \end{small}
    \caption{Samples, perturbations and histograms for MNIST digits 0 vs. 6}
    \label{fig:06}
\end{figure}

\begin{figure}
    \centering
    \begin{small}
    \begin{tabular}{rcc}
         Orig & \includegraphics[height=12mm]{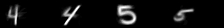} & \\ 
         Optimal & \includegraphics[height=12mm]{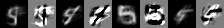} & \includegraphics[height=12mm]{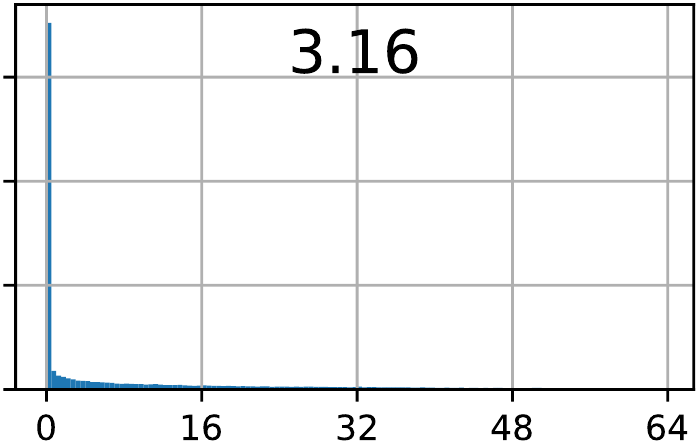}\\ 
         CNN & \includegraphics[height=12mm]{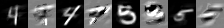} & \includegraphics[height=12mm]{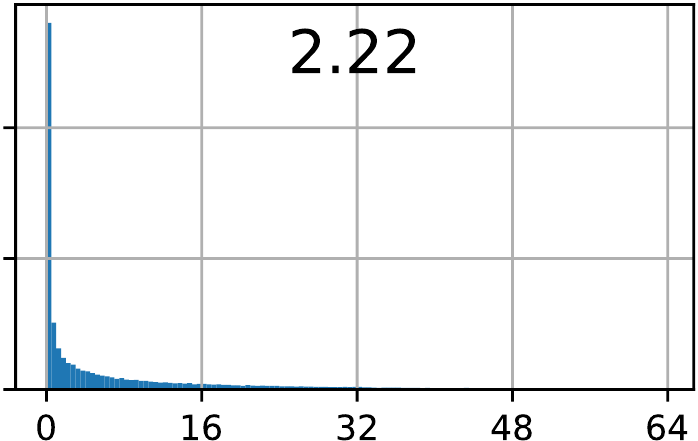}\\ 
         Lin. SVM & \includegraphics[height=12mm]{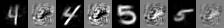} & \includegraphics[height=12mm]{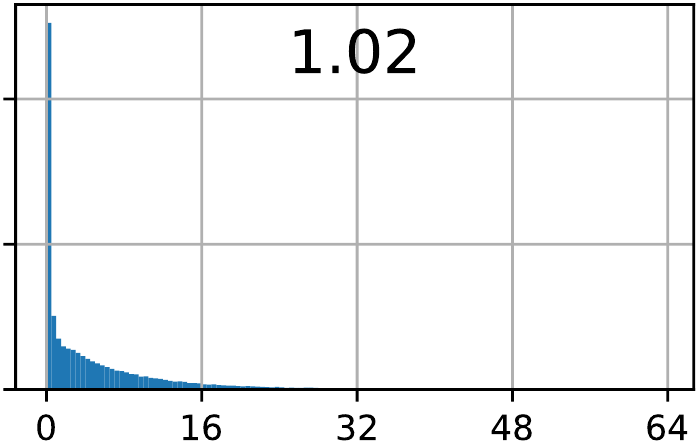}\\ 
         RBF SVM & \includegraphics[height=12mm]{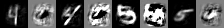} & \includegraphics[height=12mm]{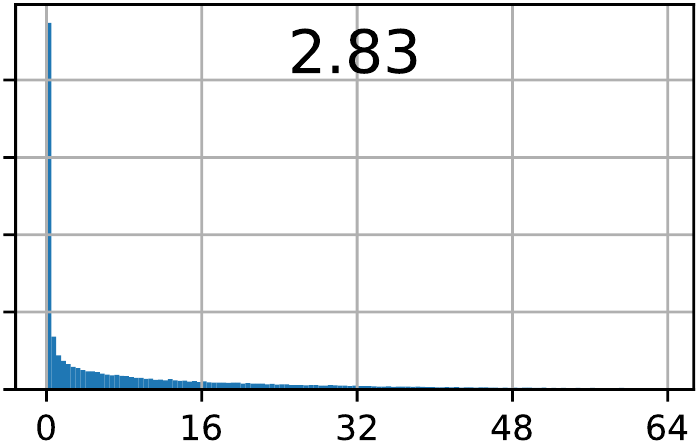}
    \end{tabular}{}
    \end{small}
    \caption{Samples, perturbations and histograms for MNIST digits 4 vs. 5}
    \label{fig:res-45}
\end{figure}

Figures~\ref{fig:res-eyeglasses}-\ref{fig:res-45} show original and adversarial samples and perturbations as well as histograms of the perturbations in pixel values for all models in different \emph{symmetric} datasets (the number at the top of each histogram is the mean perturbation L2 over all test samples).

\begin{table}
    \caption{Clean and adversarial (in brackets) classification accuracy values for different models for the \emph{symmetric} datasets.}
    \small
    \centering
    \begin{tabular}{|l|c|c|c|c|c}
\hline
Dataset / Attribute& Bayes-Optimal& CNN& Linear SVM& RBF SVM \\
\hline
CelebA / Bangs & 100\% (0\%)& 100\% (0\%)& 100\% (0\%)& 96\% (18\%) \\ 
CelebA / Black Hair & 100\% (0\%)& 98\% (2\%)& 100\% (0\%)& 96\% (26\%)\\ 
CelebA / Brown Hair & 100\% (0\%)& 99\% (1\%)& 100\% (0\%)& 86\% (16\%)\\ 
CelebA / Heavy Makeup & 100\% (0\%)& 100\% (0\%)& 100\% (0\%)& 96\% (14\%)\\ 
CelebA / High Cheekbones & 100\% (0\%)& 100\% (0\%)& 100\% (0\%)& 96\% (10\%)\\ 
CelebA / Male & 100\% (0\%)& 100\% (0\%)& 100\% (0\%)& 100\% (8\%)\\ 
CelebA / Mouth Slightly Open & 100\% (0\%)& 100\% (0\%)& 100\% (0\%)& 94\% (8\%)\\ 
CelebA / No Beard & 100\% (0\%)& 100\% (0\%)& 100\% (0\%)& 94\% (10\%)\\ 
CelebA / Smiling & 100\% (0\%)& 100\% (0\%)& 100\% (0\%)& 96\% (10\%)\\ 
CelebA / Wearing Earrings & 100\% (0\%)& 99\% (1\%)& 100\% (0\%)& 94\% (16\%)\\ 
CelebA / Wearing Lipstick & 100\% (0\%)& 100\% (0\%)& 100\% (0\%)& 92\% (18\%)\\ 
CelebA / Eyeglasses & 100\% (0\%)& 100\% (0\%)& 100\% (0\%)& 100\% (22\%)\\ 
MNIST / 06 & 100\% (0\%)& 100\% (0\%)& 100\% (0\%)& 100\% (4\%)\\ 
MNIST / 27 & 100\% (0\%)& 100\% (0\%)& 100\% (0\%)& 100\% (7\%)\\ 
MNIST / 45 & 100\% (0\%)& 100\% (0\%)& 100\% (0\%)& 100\% (8\%)\\
\hline
    \end{tabular}
    \label{tab:model-accuracy}
\end{table}{}

\subsection{Results on Real Data}

\subsubsection{Training and testing on real data}

As shown in the main paper (Figure~\ref{fig:train-on-real-data}), results on real data are consistent with our \emph{symmetric} datasets results -- CNN and Linear SVM learn a vulnerable classifier while RBF SVM is robust. Table~\ref{tab:train-on-real-data-results} lists the clean and adversarial accuracy values for the different models trained on different \emph{real} image datasets.

\begin{table*}
    \caption{Clean and adversarial (in brackets) classification accuracy values for different models for \emph{real} CelebA data.}
    \small
    \centering
    \begin{tabular}{|l|c|c|c|}
\hline
Dataset / Attribute & CNN & Linear SVM & RBF SVM \\
\hline
CelebA / Male & 94.7\% (5.30\%) & 89.7\% (10.3\%) & 89.7\% (11.3\%) \\ 
CelebA / Smiling & 88.0\% (12.0\%) & 88.3\% (11.7\%) & 84.3\% (16.3\%) \\ 
CelebA / Bangs & 94.0\% (6.00\%) & 87.0\% (13.0\%) & 93.7\% (24.7\%)\\ 
CelebA / Heavy Makeup & 86.0\% (14.0\%) & 83.7\% (16.3\%) & 85.7\% (20.7\%) \\ 
CelebA / Eyeglasses & 100.\% (0.00\%) & 93.7\% (6.33\%) & 96.3\% (32.3\%) \\
\hline
    \end{tabular}
    \label{tab:train-on-real-data-results}
\end{table*}{}

\subsubsection{Training on symmetric data and testing on real data}
To estimate how close our symmetric datasets are to the real datasets, we tested the CNNs that were trained on the symmetric dataset on real test samples and compared the test accuracy to that of CNNs that were trained on the real training data. As can be seen in Table~\ref{tab:real-data-accuracy}, there is an average accuracy reduction of just 5\%, indicating that the symmetric datasets are not that far from the original data.

\begin{table}
    \caption{Real test images classification accuracy for CNNs trained on \emph{real} training data vs CNNs trained on samples from the \emph{symmetric} dataset.}
    \small
    \centering
    \begin{tabular}{|l|c|c|}
\hline
Dataset / Attribute& Real Train Data& Symmetric Train Data \\
\hline
CelebA / Male & 92.3\% & 88.0\%\\ 
CelebA / Smiling & 89.5\% & 85.6\% \\ 
CelebA / Bangs & 94.3\% & 90.4\%  \\ 
CelebA / Heavy Makeup & 86.3\% & 79.9\% \\ 
CelebA / Eyeglasses & 98.4\% & 91.7\% \\ 
\hline
    \end{tabular}
    \label{tab:real-data-accuracy}
\end{table}{}

{}

{}

{}

{}

\clearpage
\bibliography{references}

\end{document}